%% file: main.tex
\documentclass[10pt,twocolumn,letterpaper]{article}

 \usepackage[pagenumbers]{cvpr} 

\input{preamble}
\definecolor{cvprblue}{rgb}{0.21,0.49,0.74}
\usepackage[pagebackref,breaklinks,colorlinks,allcolors=cvprblue]{hyperref}

\usepackage{multirow}
\usepackage{makecell}
\usepackage{color}

\def\paperID{} 
\def\confName{CVPR}
\def\confYear{2026}

\title{Fine-grained Image Aesthetic Assessment: \protect\\ Learning Discriminative Scores from Relative Ranks}

\author{Zhichao Yang\textsuperscript{1}\footnotemark[2] , Jianjie Wang\textsuperscript{1}\footnotemark[2] , Zhixianhe Zhang$^1$, Pangu Xie$^1$, Xiangfei Sheng$^1$, \\ Pengfei Chen$^1$, and Leida Li\textsuperscript{1,2}\footnotemark[1]\\ [-0.5mm]
\normalsize $^1$School of Artificial Intelligence, $^2$State Key Laboratory of EMIM, Xidian University \\ [-0.5mm]
\normalsize \href{https://yzc-ippl.github.io/FG-IAA/}{\normalsize \texttt{https://yzc-ippl.github.io/FG-IAA/}}
}

\begin{document}
\twocolumn[{
	\renewcommand\twocolumn[1][]{#1}
	\maketitle
	\begin{center}
		\centering
		\includegraphics[width=.93\textwidth]{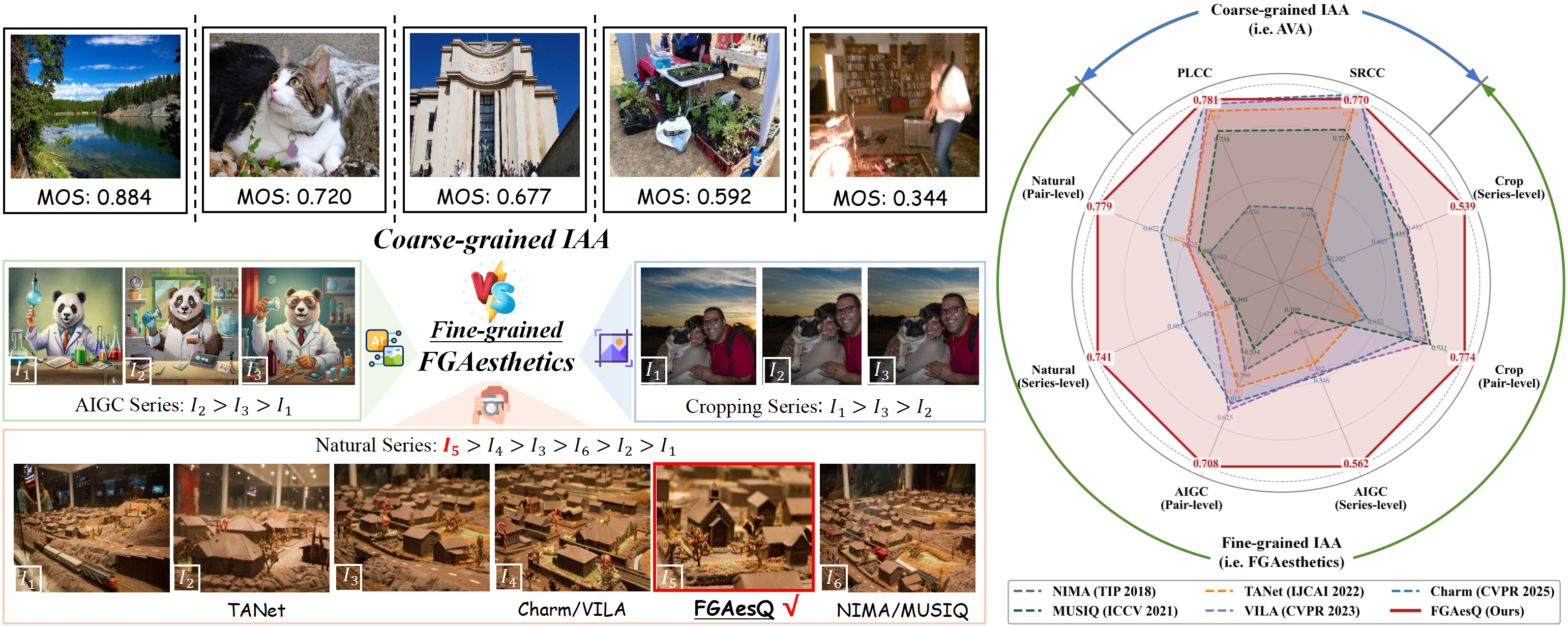}
		\captionof{figure}{\textbf{Which image is aesthetically more pleasing?} We introduce FGAesthetics, a benchmark for Fine-grained Image Aesthetic Assessment that discriminates subtle aesthetic differences among similar images in photo series. It differs from previous coarse-grained datasets, e.g., AVA, where images with notable differences are evaluated independently on an absolute scale. This enables FGAesQ, a novel IAA model, to achieve accurate aesthetic assessment in fine-grained scenarios while maintaining robust coarse-grained evaluation.}
		\label{Fig1}
	\end{center}%
}]

\renewcommand{\thefootnote}{\fnsymbol{footnote}} 
\footnotetext[2]{Equal Contribution.} 
\footnotetext[1]{Corresponding Author.}
\renewcommand{\thefootnote}{\arabic{footnote}}

\input{sec/0_abstract}    
\input{sec/1_intro}

\input{sec/2_relatedwork}
\input{sec/3_method-dataset}
\input{sec/4_method-model}

\input{sec/5_experiment}

\input{sec/6_conclusion}
\section{Acknowledgments}
This work is supported by National Natural Science Foundation of China under Grants 62471349, 62171340, 62301378, and 625B2142, Fundamental Research Funds for the Central Universities under Grant QTZX25076 and YJSJ25004, the China Postdoctoral Science Foundation under Grant 2024M762553. We also sincerely thank all annotators for their dedicated efforts and significant contributions to the construction of the database.
{
    \small
    \bibliographystyle{ieeenat_fullname}
    \bibliography{main}
}
\input{sec/X_suppl}


\end{document}

%% file: sec/0_abstract.tex
\begin{abstract}
Image aesthetic assessment (IAA) has extensive applications in content creation, album management, and recommendation systems, etc. In such applications, it is commonly needed to pick out the most aesthetically pleasing image from a series of images with subtle aesthetic variations, a topic we refer to as fine-grained IAA. Unfortunately, state-of-the-art IAA models are typically designed for coarse-grained evaluation, where images with notable aesthetic differences are evaluated independently on an absolute scale. These models are inherently limited in discriminating fine-grained aesthetic differences. To address the dilemma, we contribute \textbf{FGAesthetics}, a fine-grained IAA database with 32,217 images organized into 10,028 series, which are sourced from diverse categories including Natural, AIGC, and Cropping. Annotations are collected via pairwise comparisons within each series. We also devise Series Refinement and Rank Calibration to ensure the reliability of data and labels. Based on FGAesthetics, we further propose \textbf{FGAesQ}, a novel IAA framework that learns discriminative aesthetic scores from relative ranks through Difference-preserved Tokenization (DiffToken), Comparative Text-assisted Alignment (CTAlign), and Rank-aware Regression (RankReg). FGAesQ enables accurate aesthetic assessment in fine-grained scenarios while still maintains competitive performance in coarse-grained evaluation. Extensive experiments and comparisons demonstrate the superiority of the proposed method. 
\end{abstract}

%% file: sec/1_intro.tex
\section{Introduction}
\textit{``The difference between something good and something great is attention to detail."}
{\hfill \textit{\textemdash Charles R. Swindoll}}

\begin{table*}[ht]
	\centering
	\small
	\caption{Comparison between the proposed FGAesthetics and representative IAA datasets.}
	\begin{tabular}{lcccccl}
		\toprule
		\textbf{Dataset} & \textbf{Year} & \textbf{Num. Img} & \textbf{Data Type} & \textbf{Img Source} & \textbf{Label Type} & \textbf{Evaluation Concern} \\ 
		\midrule
		\midrule
		CUHK-PQ \cite{luo2011content} & 2011 & 17,673 & Single & Natural & Classification & Aesthetic Classification \\ 
		AVA \cite{murray2012ava} & 2012 & 255,530 & Single & Natural & MOS & Aesthetic Scoring \\ 
		AADB \cite{kong2016photo} & 2016 & 10,000 & Single & Natural & MOS & Aesthetic Attributes \\ 
		TAD66K \cite{he2022rethinking} & 2022 & 66,327 & Single & Natural/Art & MOS & Theme-specific Aesthetics \\ 
		BAID \cite{yi2023towards} & 2023 & 60,337 & Single & Art & MOS & Artistic Aesthetics \\ 
		ICAA17K \cite{he2023thinking} & 2023 & 17,726 & Single & Natural & MOS & Color Aesthetics \\ 
		AesBench \cite{huang2024aesbench} & 2024 & 2,800 & Single & Natural/AI/Art & Description & MLLM's Aesthetic Perception \\ 
		\midrule
		\midrule
		\textbf{FGAesthetics} & \textbf{2025} & \textbf{32,217} & \textbf{Series} & \textbf{Natural/AI/Crop} & \textbf{Rankings} & \textbf{Fine-grained Aesthetics} \\ 
		\bottomrule
	\end{tabular}
	\label{Tab1}
\end{table*}

Aesthetics has emerged as a pivotal focus in the creation and consumption of visual content. In view of this, image aesthetic assessment (IAA) has witnessed notable research growth \cite{deng2017image,behrad2025charm,ke2023vila,huang2024aesbench}, with extensive applications in image recommendation \cite{sun2017photo,10168279}, AI generation guidance \cite{ba2025enhancing,yang2025longt2ibench}, album management \cite{loui2003automated,datta2008algorithmic} and smart photography \cite{wu2024goal,ke2023vila}. In these scenarios, we often need to handle photo series that have the same semantic meaning but exhibit subtle aesthetic differences. This creates an urgent demand for accurate aesthetic evaluation capable of measuring and discriminating these nuanced differences, a challenge we define as \textbf{Fine-grained Image Aesthetic Assessment (FG-IAA)}.

In the past few years, numerous IAA datasets have been released, along with corresponding evaluation algorithms. Early benchmarks, such as CUHK-PQ \cite{luo2011content}, categorized images into high and low aesthetics. During this period, hand-crafted features were typically employed in aesthetic modeling \cite{datta2006studying,ke2006design}. Subsequently, deep learning methods have been widely adopted to uncover common aesthetic criteria embedded within large datasets \cite{lu2015deep,behrad2025charm}. AVA \cite{murray2012ava}, due to its extensive scale, has become the most popular IAA benchmark. Recently, there has been deeper exploration into specific aesthetic dimensions, such as color \cite{he2023thinking} and theme \cite{he2022rethinking}. The more recent prevalence of Multimodal Large Language Models (MLLMs) has also prompted researchers to explore their aesthetic perception capabilities \cite{huang2024aesbench,zhou2024uniaa}.

While existing benchmarks offer valuable insights, they typically evaluate images with notable aesthetic differences independently on an absolute scale, as illustrated in \cref{Tab1}. This inherent limitation challenges the current IAA models in fine-grained scenarios. FG-IAA presents two major challenges: (1) \textbf{Semantic Interference}: strong semantic similarities in image series hinder extracting fine-grained aesthetic differences, especially that most deep models are pre-trained for semantic tasks. (2) \textbf{Subtle Variations}: minor and diverse aesthetic differences, e.g., slight color and composition changes, require models to have robust discriminative aesthetic representations. To address these challenges, we contribute \textbf{FGAesthetics}, a benchmark specifically designed for fine-grained aesthetics, comprising 32,217 images organized into 10,028 series with aesthetic ranking labels. To ensure diversity, we collect image series from three distinct sources: Natural, AIGC, and Aesthetic cropping. These series undergo rigorous filtering via a Metrics-MLLMs-Human refinement protocol, avoiding images that are either visually significantly dissimilar or indistinguishable. Human annotators then perform pairwise comparisons within each series, where pairs with ambiguous relative ordering are filtered to calibrate the global rankings and produce the final FGAesthetics dataset.

\begin{figure*}[t]
	\centering
	\includegraphics[width=\textwidth]{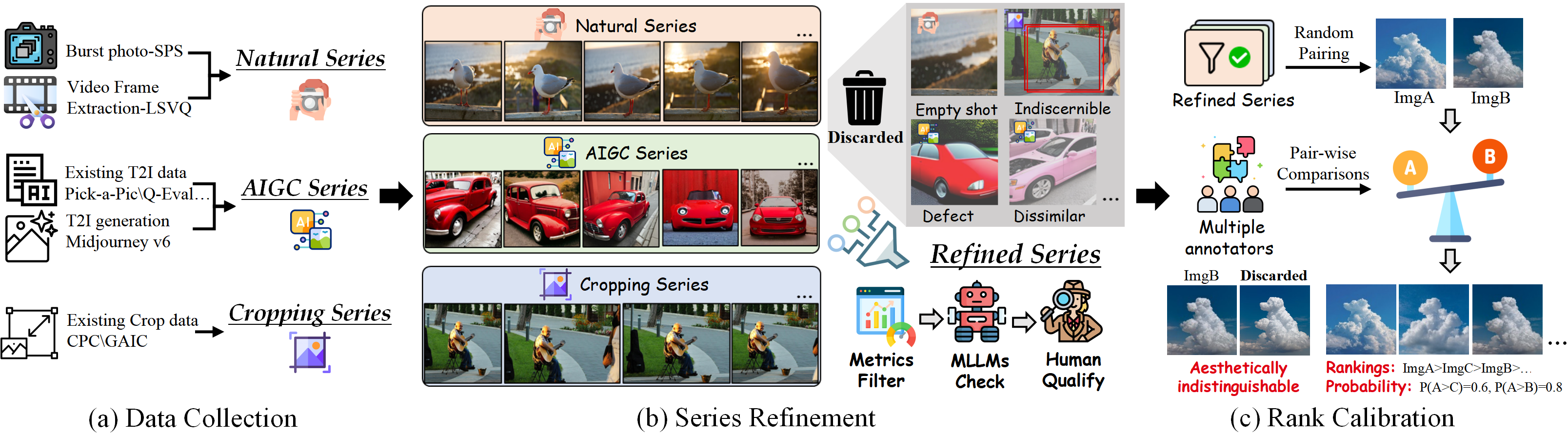}
	\caption{Overview of the construction pipeline for \textbf{FGAesthetics}. The pipeline consists of three stages: (a) Data Collection. Visually similar photo series are collected from three distinct sources: Natural, AIGC, and Cropping. (b) Series Refinement. Noisy series data undergo rigorous filtering using a Metric-MLLMs-Human refinement protocol. (c) Rank Calibration. Pairwise comparisons are annotated within each series, excluding data that cannot be aesthetically distinguished to obtain calibrated aesthetic rankings.}
	\label{Fig2}
\end{figure*}

Building on FGAesthetics, we further propose \textbf{FGAesQ}, a new IAA model that enables accurate aesthetic assessment in fine-grained scenarios while maintaining robust performance in coarse-grained evaluation (\cref{Fig1}-right). This is achieved by learning  discriminative aesthetic scores from relative ranks, where coarse-grained data establishes foundational aesthetic perception and fine-grained data refines the regression space for subtle aesthetic distinctions. Specifically, we introduce Difference-preserved Tokenization (DiffToken) that maintains distinctive details between similar images while scaling homogeneous regions. Comparative Text-assisted Alignment (CTAlign) is designed to further enhance the discrimination of visual representations. Finally, we develop Rank-aware Regression (RankReg), which leverages aesthetic rankings to calibrate score predictions, ensuring consistency between absolute assessments and relative aesthetic ordering. 

The contributions of this work are three-fold:
\begin{itemize}
	\item We define Fine-grained Image Aesthetic Assessment and contribute FGAesthetics comprising 32,217 images from diverse sources including Natural, AIGC, and Cropping, organized into 10,028 series with aesthetic rankings.
	\item We propose FGAesQ, a novel IAA model that learns discriminative scores from relative ranks, achieving accurate aesthetic assessment in fine-grained scenarios while maintaining robust coarse-grained evaluation.
	\item Through extensive experiments and comparative analysis, we reveal fundamental limitations of current IAA methods in capturing fine-grained aesthetic differences and demonstrate the superiority of FGAesQ.
\end{itemize}

%% file: sec/2_relatedwork.tex
\section{Related Work}
\subsection{Image Aesthetic Assessment Datasets} 
In the past few years, a proliferation of datasets have been developed for image aesthetic assessment, with representative benchmarks summarized in \cref{Tab1}. Early datasets are designed for different aesthetic tasks, with CUHK-PQ \cite{luo2011content} for binary classification and AVA \cite{murray2012ava} for score regression, where AVA remains the most widely used benchmark due to its large scale. Then, specific aesthetic dimensions have been explored: AADB \cite{kong2016photo} targets aesthetic attributes, e.g., lighting and depth of field; TAD66K \cite{he2022rethinking} focuses on theme-specific aesthetics; BAID \cite{yi2023towards} addresses artistic aesthetics; and ICAA17K \cite{he2023thinking} emphasizes color aesthetics. With the emergence of Multimodal Large Language Models (MLLMs), benchmarks like AesBench \cite{huang2024aesbench} have been developed to evaluate their aesthetic perception capabilities.

The above benchmarks focus on the holistic aesthetic assessment of individual images, making them unsuitable for fine-grained evaluation scenarios, while related datasets have not been thoroughly studied due to open-source limitations. Chang \textit{et al.} \cite{chang2016automatic} proposed the Series Photo Selection task for burst photography scenarios, but the partially open-sourced dataset limits its development. Ren \textit{et al.} \cite{ren2020best} introduced the innovative Best Frame Selection task to identify the most aesthetically pleasing frames from short videos, yet the data remains inaccessible. In this study, we define the task of fine-grained image aesthetic assessment, and contribute FGAesthetics with diverse data types including Natural, AIGC, and Cropping, addressing the gap for fine-grained evaluation scenarios.

\subsection{Image Aesthetic Assessment Models}
Early IAA models relied on hand-crafted features and simple machine learning techniques, mainly drawing inspirations from intuitive judgments of image aesthetics and widely accepted photography rules \cite{datta2006studying,ke2006design}. With the advent of deep learning, IAA methods began leveraging deep neural networks to identify common aesthetic criteria in large datasets. NIMA \cite{talebi2018nima} used EMD-based loss to train deep models for aesthetic distribution prediction. TANet \cite{he2022rethinking} and TMCR \cite{yang2024semantics} explored the intricate coupling of aesthetics and themes. VILA \cite{ke2023vila} and AesCLIP \cite{sheng2023aesclip} achieved aesthetic evaluation by leveraging user-generated aesthetic reviews. MUSIQ \cite{ke2021musiq} and Charm \cite{behrad2025charm} address the fixed shape constraint of Vision Transformers (ViTs) architectures, enabling variable-sized inputs that preserve compositional integrity for more accurate aesthetics modeling. Recently, MLLM-based aesthetic scorers have been explored, with Q-Align \cite{wu2024q} and UNIAA \cite{zhou2024uniaa} using discrete level labels to guide MLLMs in aesthetic score regression. While these models have achieved significant progress in coarse-grained aesthetic evaluation for images with notable aesthetic differences on an absolute scale, their effectiveness on fine-grained scenarios remains largely unexplored. In this study, we address this and propose FGAesQ, a novel IAA framework with superior performance on fine-grained scenarios while still maintains robust coarse-grained evaluation.

%% file: sec/3_method-dataset.tex
\section{FGAesthetics}
In this section, we detail FGAesthetics, with the construction pipeline illustrated in \cref{Fig2}. Throughout the benchmark, we design multiple strategies to ensure diversity and reliability. We will elaborate on the construction process from three aspects: data collection, series refinement, and rank calibration, followed by a statistical analysis.

\subsection{Data Collection}
To ensure diversity and comprehensive evaluation of fine-grained aesthetics, we collect images from diverse sources. For natural image series, we sample from two sources: burst photographs capturing identical scenes from the SPS \cite{chang2016automatic}, and sequential frames extracted from videos in the LSVQ \cite{ying2021patch}. For AI-generated content (AIGC) series, we primarily leverage text-to-image data where images generated from same text prompts form a series. This data is mainly sampled from established T2I evaluation benchmarks, including Pick-a-pic \cite{kirstain2023pick}, Q-Eval-100K \cite{zhang2025q}, and NIGHTS \cite{fu2023dreamsim}. We also utilize advanced proprietary generative models, such as Midjourney v6 \cite{midjourney}, to generate series by varying parameters. Additionally, previous IAA research has demonstrated the significant impact of varying resolutions, particularly aspect ratio, on aesthetic modeling \cite{ke2021musiq,behrad2025charm}. To incorporate this factor in fine-grained IAA, we include cropping series derived from datasets containing dense cropping variations of the same image, such as CPC \cite{wei2018good} and GAIC \cite{zeng2019reliable}. We ensured balanced sampling across all three categories, with detailed source distribution maps provided in the Appendix.

\subsection{Series Refinement}
To ensure that series data contains visually similar yet distinguishable images, we design a Metrics-MLLMs-Human refinement protocol to filter each collection. For Metrics Filtering, both generic and domain-specific measures are utilized to screen the data. Specifically, SSIM \cite{wang2004image} quantifies similarity between images within each series, with samples below threshold values being removed. For cropping, IoU \cite{rezatofighi2019generalized} is utilized to prevent indistinguishable cropping frames. For AIGC series, T2I alignment scores identify and eliminate obviously dissimilar images. Given the powerful contextual understanding capabilities of MLLMs \cite{tai2024link}, we also employ Gemini-2.5-pro \cite{team2023gemini} to validate the filtered series by providing all images within each series and prompting it to check data based on visual similarity and contextual coherence. Finally, all series undergo qualification by five human annotators to ensure data quality. This rigorous multi-stage process yields high-quality fine-grained series data for the subsequent aesthetic ranking annotation. Implementation details of metrics filtering, the instructions for MLLMs check using Gemini, and the operational interface for human qualify are included in the Appendix.

\subsection{Rank Calibration}
Based on the refined data, we perform pairwise comparisons within each series to obtain global aesthetic rankings. Specifically, for a series of length $n$, a total of $C_{n}^{2}$ image pairs are generated. Each pair is evaluated by 10 trained annotators for aesthetic comparison, determining which image is superior or indistinguishable. This annotation methodology offers three key advantages: (1) Pairwise aesthetic comparisons yield more reliable annotations compared to absolute scoring or full series ranking. (2) It facilitates the identification and removal of aesthetically indistinguishable samples. (3) The probability of pairwise comparisons becomes directly measurable, e.g., 0.5 indicates no consensus (half preferring each image) and warrants filtering while 0.9 suggests clear aesthetic superiority. Based on the collected pairwise annotations, we derive the global ranking labels, while filtering out indistinguishable samples and series with comparison conflicts. Note that for sampled data from SPS \cite{chang2016automatic}, originally annotated through pairwise comparisons, fine-grained rankings can be directly derived without additional human annotation.

\begin{figure}[t]
	\centering
	\includegraphics[width=0.47\textwidth]{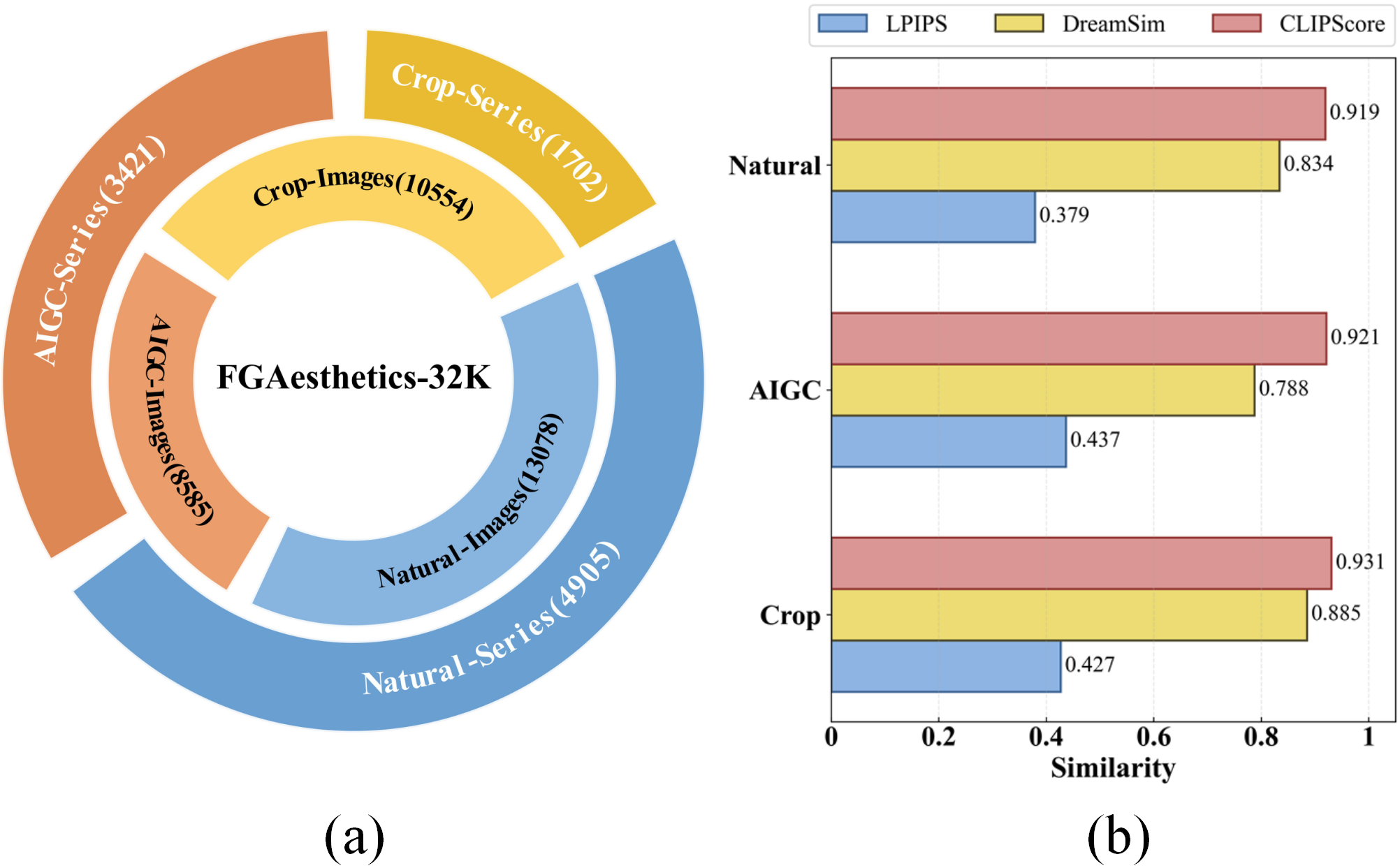}
	\caption{Statistical analysis of FGAesthetics. (a) Image and series count distribution across Natural, AIGC, and Cropping. (b) Within-series similarity distributions measured by LPIPS (low-level) \cite{zhang2018unreasonable}, DreamSim (mid-level) \cite{fu2023dreamsim}, and CLIPScore (high-level) \cite{hessel2021clipscore} across three sources. Note that LPIPS and DreamSim are subtracted from 1, ensuring consistent polarity with CLIPScore, where higher values indicate greater similarity.}
	\label{Fig3}
\end{figure}

\subsection{Statistics of FGAesthetics}
During data collection, we gather 106,632 images, which are filtered to 32,588 through rigorous series refinement. Following rank calibration, the final dataset comprises 32,217 images organized into 10,028 series with lengths ranging from 2 to 10, forming FGAesthetics. \cref{Fig3}-(a) presents image and series count distribution across Natural, AIGC, and Cropping. The analysis shows relatively balanced image distribution, with Cropping containing more long series, as multiple crops from the same source image naturally form longer sequences with finer aesthetic distinctions. To quantitatively verify the fine-grained nature of our dataset, we measure within-series similarity using LPIPS (low-level) \cite{zhang2018unreasonable}, DreamSim (mid-level) \cite{fu2023dreamsim}, and CLIPScore (high-level) \cite{hessel2021clipscore}, as summarized in \cref{Fig3}-(b). Results reveal consistent trends across three sources: high semantic similarity (CLIPScore$>$$0.91$), contrasted with moderate patch-level similarity (LPIPS between $0.379$ and $0.437$). This confirms that images within series maintain semantic coherence while exhibiting fine-grained visual differences that influence aesthetic perception.

\begin{figure*}[t]
	\centering
	\includegraphics[width=0.97\textwidth]{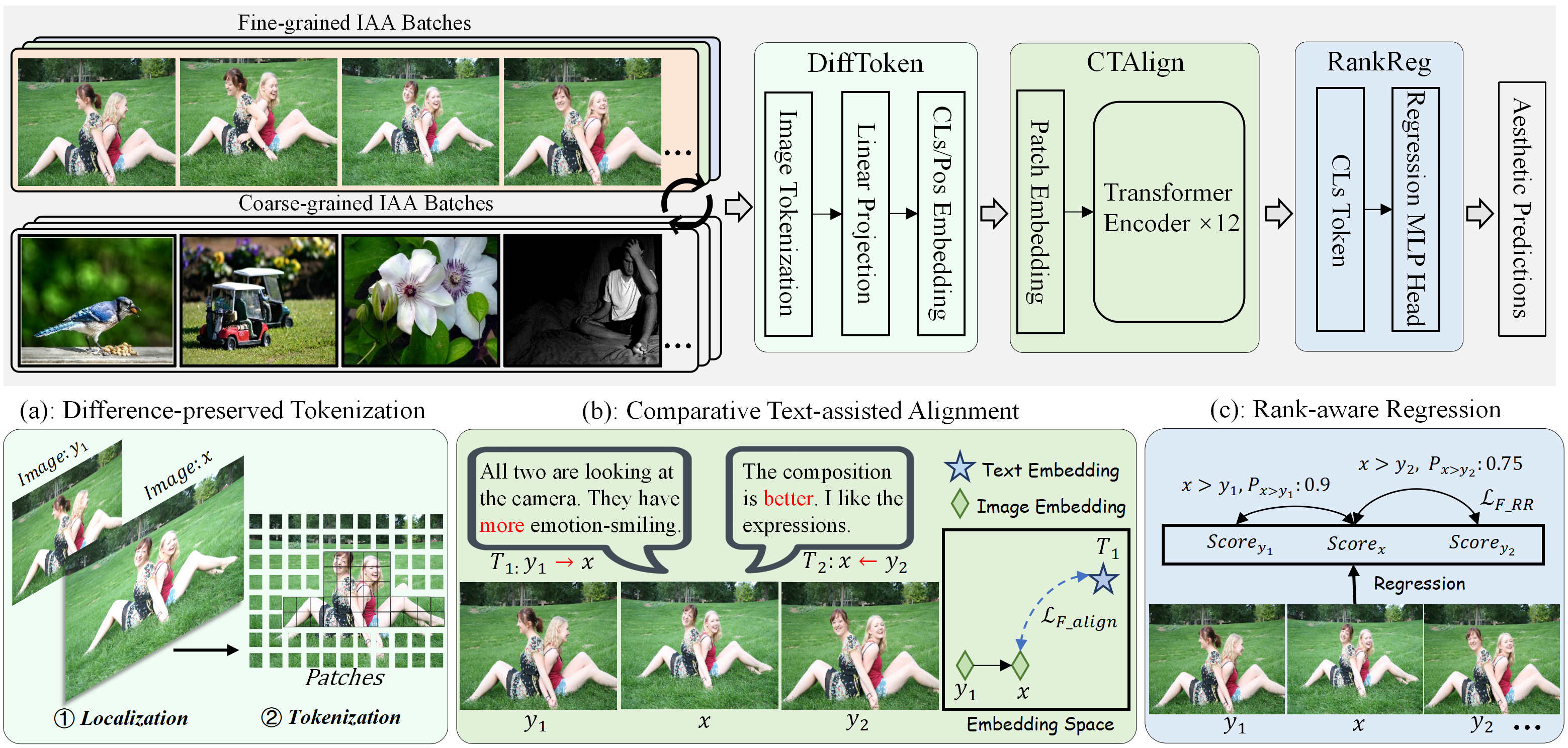}
	\caption{Overall pipeline of the proposed \textbf{FGAesQ}. FGAesQ learns discriminative aesthetic scores from relative ranks through three modules: (a) Difference-preserved Tokenization (DiffToken) selectively maintains difference regions at their original resolution while downscaling others. (b) Comparative Text-assisted Alignment (CTAlign) achieves distinctive aesthetic visual representations. (c) Rank-aware Regression (RankReg) rectifies the coarse-grained score regression with fine-grained aesthetic rankings.}
	\label{Fig4}
\end{figure*}

%% file: sec/4_method-model.tex
\section{FGAesQ}
Building upon FGAesthetics, we propose FGAesQ, a novel IAA framework that enables accurate aesthetic assessment in fine-grained scenarios while maintaining robust coarse-grained evaluation. The overall pipeline is shown in \cref{Fig4}.

\noindent \textbf{Overall architecture.} Fine-grained aesthetic discrimination is fundamentally grounded on coarse-grained aesthetic perception. Therefore, we propose to leverage fine-grained ranks to enhance scoring discriminability, enabling unified capabilities for both coarse- and fine-grained aesthetic evaluations. Specifically, we train a Vision Transformer \cite{radford2021learning} using both coarse- and fine-grained aesthetic data: the coarse batch contains distinct images with absolute scores from existing IAA datasets, while the fine batch contains visually similar image series with ranking labels. To address the Semantic Interference and Subtle Variations challenges in FG-IAA, we introduce difference-preserved tokenization (DiffToken), comparative text-assisted alignment (CTAlign) and rank-aware regression (RankReg), based on which discriminative aesthetic representations can be learned.

\subsection{Difference-preserved Tokenization}
For fine-grained input, images within a series share substantial semantic similarity, with only limited regions exhibiting difference significantly influencing their aesthetic ranking. This characteristic is empirically supported by the low perceptual similarity measured by LPIPS (operating on $64\times64$ patches) shown in \cref{Fig3}-(b). Recognizing that these aesthetics-decisive regions contain critical cues, we design Difference-preserved Tokenization (DiffToken), which maintains their details while downscaling others. This approach enables the visual model to focus computational resources on perceptually significant areas that drive fine-grained aesthetic discrimination.

DiffToken first localizes aesthetics-decisive regions and then performs mix-resolution tokenization. Specifically, for a fine-grained batch containing $k$ series: $ B_{F}\left \{ s_{1},...s_{k}\right \} $, we randomly sample an image pair $ \left ( x,y_{1}  \right ) $ from a series, where $x$ is the target image to be tokenized and $ y_{1}$ is the reference image. Using the longer edge of $x$ as the baseline, $ y_{1}$ is aligned through scaling while maintaining its original aspect ratio. Both images are then partitioned into patches at a multiple (e.g., 2×) of the standard ViT patch size. Based on this partition, the similarity between corresponding patches is computed using RGB-channel SSIM \cite{wang2004image}:
\begin{equation}
	s_{i,j} ={\rm SSIM}\left ( P_{i,j}^{x}, P_{i,j}^{y_{1} } \right ),
\end{equation}
where $P_{i,j}^{x}, P_{i,j}^{y_{1}}$ denote patches at position $ \left ( i,j \right ) $ in images $x$ and $ y_{1}$ respectively. We then sort all patches by similarity scores $ s_{i,j}$ and identify the aesthetics-decisive regions by selecting patches with similarity below a threshold $ \tau $:
	\begin{equation}
		D=\left \{ \left ( i,j \right )|s_{i,j}<\tau, \tau={\rm percentile}(s,p)  \right \},
	\end{equation}
where $p$ is the percentile parameter controlling the proportion of patches identified as aesthetics-decisive regions.

Based on identified aesthetics-decisive regions, we implement difference-preserved tokenization. Patches within set $D$ are partitioned at the original ViT patch resolution to preserve details. Remaining patches are scaled to required size with random dropping applied to satisfy token count constraints. Additionally, we accommodate varying input aspect ratios through bilinear interpolation of position embeddings during the tokenization process \cite{havtorn2023msvit,ronen2023vision,wang2023exploring,ke2021musiq,behrad2025charm}. This ensures both high-fidelity local difference and comprehensive global composition.

\subsection{Comparative Text-assisted Alignment}
Based on difference-preserved tokens, we further utilize comparative textual descriptions to orient the visual model toward fine-grained aesthetic differences, thereby enhancing discrimination capabilities. To obtain these descriptions while minimizing human annotation burden, we leverage the robust reasoning capabilities of MLLMs. Specifically, image pairs with human-annotated aesthetic ranking labels are utilized to instruct GPT-4o \cite{achiam2023gpt}, eliciting comparative reasoning with explicitly contrastive vocabulary. For Cropping and AIGC series, original images and text prompts also serve as context to ensure high-quality textual outputs. The detailed prompt engineering for reasoning generation is provided in the Appendix.

Given $T_{1}:x\gets y_{1}$ representing the comparative text of target image $x$ relative to reference $ y_{1}$, we extract its embedding using text encoder of CLIP \cite{radford2021learning} and align it with the visual embedding difference between the two images. This alignment is achieved by minimizing the following loss:
\begin{equation}
	\mathcal{L}_{F\_align} = \cos(E_v(x) - E_v(y_1), E_t(T_1)),
\end{equation}
where $ E_v $, $ E_t $ denote the vision and text encoders respectively, and $\cos()$ is the cosine similarity. Note that texts are only utilized during the training process, and only the learned image encoder is used during inference.

\subsection{Rank-aware Regression}
With the aligned visual representations, human-annotated aesthetic ranking labels within each series are utilized to refine score regression. Specifically, for an image pair $ \left ( x,y_{1}  \right ) $ from a series, absolute scores $\left ( Score_{x}, Score_{y_{1}}  \right ) $ are first obtained through a regression head. We then employ the Bradley-Terry model \cite{bradley1952rank} to compute the probability that target image $x$ is aesthetically superior to reference $ y_{1}$:
\begin{equation}
	P_{(x \succ y_1)} = \frac{e^{Score_x}}{e^{Score_x} + e^{Score_{y_1}}}.
\end{equation}
The above operation is performed for all pairs within the series to obtain the complete predicted probability distribution $ \mathbf{P'}=\left \{ P_{(x \succ y_1)},P_{(x \succ y_2)},... \right \} $. The ListMLE \cite{xia2008listwise} is used as loss function to align it with the ground truths $\mathbf{P}$:
\begin{equation}
	\mathcal{L}_{F\_RR}( \mathbf{P'}, \mathbf{P}) = -\sum_{j=1}^{n} \log \frac{e^{\mathbf{P'}(r_i)}}{\sum_{j=i}^{n} e^{\mathbf{P'}(r_j)}},
\end{equation}
where $r$ represents the ranking order derived from $\mathbf{P}$, $\mathbf{P'}(r_i)$ is the predicted probability at position $i$ in this ranking, and $n$ is the cardinality of $\mathbf{P'}$. The loss $\mathcal{L}_{F\_RR}$ simultaneously optimizes both the ranking accuracy of series elements and the calibration of probability estimates between pairs.

\subsection{Training of FGAesQ}
FGAesQ adopts a two-stage training strategy: pre-training on coarse-grained data followed by joint learning across both granularities. The model is first pre-trained using coarse-grained IAA data to establish foundational aesthetic perception capability. Similar to previous IAA methods \cite{talebi2018nima,he2022rethinking,ke2023vila,yang2024semantics,behrad2025charm,yang2025language,sheng2025fine,sheng2025tuningiqa}, the Earth Mover's Distance (EMD) loss is employed to guide the model in predicting aesthetic scores for single images, denoted as $\mathcal{L}_{C\_EMD}$. Building upon this, we perform a joint learning that alternately optimizes the model using coarse- and fine-grained IAA batches, enabling fine-grained ranking to refine coarse-grained scoring. This alternating training process can be formatted as:
\begin{equation}
	\mathcal{L} =  \underbrace{ \delta \cdot  (\lambda\mathcal{L}_{F\_align}+ \mathcal{L}_{F\_RR})}_{\text{Fine-grained}}+\underbrace{\left ( 1-\delta    \right ) \cdot \mathcal{L}_{C\_EMD}}_{\text{Coarse-grained}},
\end{equation}
where $\delta$ represents a binary alternating indicator, and $\lambda$ is a balancing coefficient. The optimization of FGAesQ proceeds through momentum-based updates, ensuring smooth optimization across different learning objectives.

%% file: sec/5_experiment.tex
\section{Experiments}

\begin{table*}[t]
	\centering
	\caption{Performance comparison between the proposed FGAesQ and state-of-the-art IAA methods on FGAesthetics. Results are reported across all three image sources (Natural, AIGC, and Cropping) for both pair-level and series-level evaluations. Model parameter counts are also provided. The top two results are highlighted in \textbf{bold} and \underline{underlined}, respectively.}
	\resizebox{\linewidth}{!}{
		\begin{tabular}{wl{3.8cm}|c|*{2}{wc{0.8cm}}|*{2}{wc{0.8cm}}|*{2}{wc{0.8cm}}|*{2}{wc{0.8cm}}|*{2}{wc{0.8cm}}|*{2}{wc{0.8cm}}}
			\toprule
			\multirow{3}{*}{\textbf{Methods}} & \multirow{3}{*}{\textbf{Params}} & \multicolumn{4}{c|}{\textbf{Natural}} & \multicolumn{4}{c|}{\textbf{AIGC}} & \multicolumn{4}{c}{\textbf{Cropping}} \\
			\cline{3-14}\rule{0pt}{12pt}
			& & \multicolumn{2}{c|}{\textit{Pair-level}} & \multicolumn{2}{c|}{\textit{Series-level}} & \multicolumn{2}{c|}{\textit{Pair-level}} & \multicolumn{2}{c|}{\textit{Series-level}} & \multicolumn{2}{c|}{\textit{Pair-level}} & \multicolumn{2}{c}{\textit{Series-level}} \\ 
			\cline{3-14}\rule{0pt}{12pt}
			& & \textit{Acc} & \textit{F1} & \textit{s-Acc} & \textit{s-SRCC} & \textit{Acc} & \textit{F1} & \textit{s-Acc} & \textit{s-SRCC} & \textit{Acc} & \textit{F1} & \textit{s-Acc} & \textit{s-SRCC} \\ 
			\midrule
			\midrule
			NIMA (TIP 2018) \cite{talebi2018nima}    & 54.3M & 0.589 & 0.588 & 0.493 & 0.225 & 0.566 & 0.565 & 0.379 & 0.137 & 0.655 & 0.649 & 0.272 & 0.312 \\
			MLSP (CVPR 2019) \cite{hosu2019effective}    & 114 M & 0.608 & 0.607 & 0.531 & 0.255 & 0.604 & 0.602 & 0.414 & 0.182 & 0.710 & 0.710 & 0.396 & 0.430 \\
			MUSIQ (ICCV 2021) \cite{ke2021musiq}  & 78.6M & 0.607 & 0.606 & 0.487 & 0.233 & 0.535 & 0.532 & 0.344 & 0.054 & 0.731 & 0.731 & 0.375 & 0.495 \\
			TANet (IJCAI 2022) \cite{he2022rethinking} & 13.9M  & 0.629 & 0.628 & 0.539 & 0.281 & 0.591 & 0.590 & 0.447 & 0.195 & 0.641 & 0.641 & 0.244 & 0.295 \\
			VILA (CVPR 2023) \cite{ke2023vila}   & 303M  & 0.625 & 0.625 & 0.546 & 0.296 & 0.625 & 0.624 & 0.438 & 0.245 & 0.726 & 0.726 & 0.423 & 0.443 \\
			Charm (CVPR 2025) \cite{behrad2025charm}  & 85.7M  & 0.672 & 0.672 & 0.602 & 0.404 & 0.616 & 0.614 & 0.448 & 0.247 & 0.707 & 0.707 & 0.381 & 0.432 \\ 
			\midrule
			\midrule
			Q-Align (ICML 2024) \cite{wu2024q}            &8.20B  & 0.711 & 0.711 & 0.676 & 0.496 & 0.646 & 0.646 & 0.462 & 0.310 & 0.738 & 0.739 & 0.413 & 0.487 \\
			UNIAA (arXiv 2024)  \cite{zhou2024uniaa}           & 7.06B & 0.709 & 0.708 & 0.656 & 0.486 & 0.640 & 0.639 & 0.486 & 0.256 & 0.675 & 0.670 & 0.300 & 0.386 \\
			RealQA (arXiv 2025) \cite{li2025next}    & 8.29B & 0.658 & 0.666 & 0.593 & 0.382 & 0.656 & 0.656 & 0.464 & 0.308 & 0.605 & 0.613 & 0.275 & 0.235 \\
			\midrule
			\midrule
			NIMA (Fine-tuned) \cite{talebi2018nima}              & 54.3M & 0.643 & 0.644 & 0.536 & 0.312 & 0.598 & 0.597 & 0.429 & 0.196 & 0.708 & 0.704 & 0.277 & 0.352 \\
			MUSIQ (Fine-tuned) \cite{ke2021musiq}              & 78.6M & 0.654 & 0.654 & 0.545 & 0.356 & 0.572 & 0.568 & 0.386 & 0.145 & \underline{0.770} & \underline{0.770} & \underline{0.487} & \underline{0.556} \\
			VILA (Fine-tuned) \cite{ke2023vila}              &303M  & 0.693 & 0.692 & 0.643 & 0.430 & 0.650 & 0.647 & 0.504 & 0.291 & 0.714 & 0.710 & 0.373 & 0.439 \\
			Charm (Fine-tuned) \cite{behrad2025charm}             &85.7M  & 0.723 & 0.723 & 0.670 & 0.474 & 0.620 & 0.619 & 0.478 & 0.254 & 0.755 & 0.755 & 0.467 & 0.517 \\ 
			\midrule
			\midrule
			\textbf{FGAesQ} (w/o DiffToken) &86.3M & \underline{0.773} & \underline{0.773} & \underline{0.720} & \underline{0.664} & \underline{0.688} & \underline{0.686} & \underline{0.517} & \underline{0.442} & 0.764 & 0.763 & 0.483 & 0.537 \\
			\textbf{FGAesQ} (w DiffToken)          &86.3M & \textbf{0.779} & \textbf{0.779} & \textbf{0.753} & \textbf{0.729} & \textbf{0.709} & \textbf{0.707} & \textbf{0.561} & \textbf{0.482} &    \textbf{0.774} & \textbf{0.773} & \textbf{0.488}  & \textbf{0.590} \\ 
			\bottomrule
	\end{tabular}}
	\label{Tab2}
\end{table*}

\subsection{Experimental Settings}

\textbf{Implementation Details.}
For FGAesQ, we utilize visual encoder (ViT-B/16) of CLIP \cite{radford2021learning} as the backbone, where AVA \cite{murray2012ava} provides coarse-grained data and FGAesthetics enables fine-grained training. For coarse-grained pre-training, we use a learning rate of 2e-5, batch size of 128, and train for 3 epochs. For joint learning, we maintain a learning rate of 2e-5 with weight decay of 5e-5, using batch sizes of approximately 64 for fine-grained data (with minor variations due to series length) and 128 for coarse-grained, with $\lambda$ set to 10. The momentum update coefficients are set to 0.615 and 0.8 respectively, with training continuing for 7 epochs on an A800 GPU. For the DiffToken module, we employ 32×32 patches for difference localization with $p$ set to 0.5, further details on these settings are in the Appendix.

\noindent \textbf{Evaluation Setting.}
We employ an 8:1:1 train-validation-test split for FGAesthetics, ensuring balanced distribution of all three image sources. For fine-grained evaluation, we conduct pair-level and series-level testings. At pair-level, we use \textit{Acc} and \textit{F1} as metrics to assess local discrimination within series. For series-level, we measure the accuracy of identifying the most aesthetic image and use SRCC for ranking correlation.  Considering longer series pose greater challenges, we employ length-weighted metrics, denoted as \textit{s-Acc} and \textit{s-SRCC}. For coarse-grained evaluation, we adopt the standard SRCC and PLCC. Competing methods contain CNN-based, ViT-based, and MLLM-based IAA models.

\subsection{Performance Evaluation}

\textbf{Benchmark Results on FGAesthetics.}
\cref{Tab2} lists the comparative results of FGAesQ against existing IAA methods on FGAesthetics, from which we have several key observations. First, existing IAA models demonstrate significant performance degradation on fine-grained evaluation scenarios, particularly for series-level testing. Second, MLLM-based IAA approaches outperform traditional deep learning-based methods, with Q-Align \cite{wu2024q} showing very encouraging results even against fine-tuned models. This stems from the enhanced fine-grained perception capabilities afforded by larger parameters. Third, input scale-focused methods \cite{hosu2019effective, ke2021musiq} exhibit superior performance on cropping data, with fine-tuned MUSIQ \cite{ke2021musiq} achieving second-best results. Finally, FGAesQ achieves the best performance across all evaluation protocols, demonstrating superior aesthetic discrimination capabilities in FG-IAA. Additionally, we evaluate FGAesQ excluding DiffToken during inference, which operates without reference image comparison yet still maintains competitive performance.

\begin{table}[t]
	\centering
	\caption{Balance between coarse-grained (AVA) and fine-grained (FGAesthetics) evaluations. Metrics shown are (\textit{Acc}, \textit{F1})/2 for Pair and (\textit{s-Acc}, \textit{s-SRCC})/2 for Series. Best in bold.}
	\resizebox{\linewidth}{!}{
		\begin{tabular}{wc{3.5cm}|cc|cc}
			\toprule
			\multirow{2}{*}{\textbf{Methods}} & \multicolumn{2}{c|}{\textbf{AVA}} & \multicolumn{2}{c}{\textbf{FGAesthetics}} \\
			\cline{2-5}\rule{0pt}{12pt}
			& SRCC & PLCC & Pair & Series \\
			\midrule
			\midrule
			NIMA (TIP 2018) \cite{talebi2018nima}  & 0.633 & 0.647 & 0.602 & 0.303  \\
			MUSIQ (ICCV 2021) \cite{ke2021musiq} & 0.726 & 0.738 & 0.624 & 0.331  \\
			VILA (CVPR 2023) \cite{ke2023vila}  & 0.774 & 0.774 & 0.659 & 0.399  \\
			Charm (CVPR 2025) \cite{behrad2025charm} & \textbf{0.777} & 0.779 & 0.665 & 0.419 \\
			\midrule
			NIMA (Fine-tuned) \cite{talebi2018nima} & \makecell{0.330\\\textcolor{red}{(\small -0.303)}} & \makecell{0.313\\\textcolor{red}{(\small -0.334)}} & \makecell{0.649\\\textcolor[RGB]{14, 127, 63}{(\small +0.047)}} & \makecell{0.350\\\textcolor[RGB]{14, 127, 63}{(\small +0.047)}} \\
			MUSIQ (Fine-tuned) \cite{ke2021musiq}  & \makecell{0.437\\\textcolor{red}{(\small -0.289)}} & \makecell{0.461\\\textcolor{red}{(\small -0.277)}} & \makecell{0.665\\\textcolor[RGB]{14, 127, 63}{(\small +0.041)}} & \makecell{0.413\\\textcolor[RGB]{14, 127, 63}{(\small +0.082)}} \\
			VILA (Fine-tuned) \cite{ke2023vila} & \makecell{0.522\\\textcolor{red}{(\small -0.252)}} & \makecell{0.507\\\textcolor{red}{(\small -0.267)}} & \makecell{0.687\\\textcolor[RGB]{14, 127, 63}{(\small +0.028)}} & \makecell{0.447\\\textcolor[RGB]{14, 127, 63}{(\small +0.048)}} \\
			Charm (Fine-tuned) \cite{behrad2025charm} & \makecell{0.470\\\textcolor{red}{(\small -0.307)}} & \makecell{0.472\\\textcolor{red}{(\small -0.307)}} & \makecell{0.699\\\textcolor[RGB]{14, 127, 63}{(\small +0.034)}} & \makecell{0.477\\\textcolor[RGB]{14, 127, 63}{(\small +0.058)}} \\
			\midrule
			\midrule
			\textbf{FGAesQ}           & 0.770 & \textbf{0.781} & \textbf{0.753} & \textbf{0.600} \\
			\bottomrule
	\end{tabular}}
	\label{Tab3}
\end{table}

\noindent \textbf{Balance between Coarse- and Fine-grained IAA.}
A robust IAA model is expected to handle both coarse- and fine-grained evaluation scenarios. \cref{Tab3} compares performance on both tasks, where FGAesQ achieves the highest average results. Existing SOTA models fine-tuned with ranking loss show enhanced fine-grained capabilities but suffer severe coarse-grained degradation (SRCC and PLCC drops highlighted in red). In contrast, FGAesQ maintains competitive performance across both tasks, demonstrating our approach effectively learns discriminative scores from relative ranks.

\begin{table}[t]
	\centering
	\caption{Cross-dataset validation on three IAA benchmarks: ICAA17K \cite{he2023thinking}, AADB \cite{kong2016photo}, and TAD66K \cite{he2022rethinking}. Results for UNIAA \cite{zhou2024uniaa} on AADB are omitted as data is used in training. }
	\resizebox{\linewidth}{!}{
			\begin{tabular}{c|cc|cc|cc}
					\toprule
					\multirow{2}{*}{\textbf{Methods}} & \multicolumn{2}{c|}{\textbf{ICAA17K}} & \multicolumn{2}{c|}{\textbf{AADB}} & \multicolumn{2}{c}{\textbf{TAD66K}} \\
					\cline{2-7}
					\rule{0pt}{12pt}
					& SRCC & PLCC & SRCC & PLCC & SRCC & PLCC \\
					\midrule
					\midrule
					NIMA \cite{talebi2018nima}   & 0.629 & 0.629 & 0.365 & 0.371 & 0.392 & 0.370 \\
					MUSIQ \cite{ke2021musiq} & 0.581 & 0.578 & 0.331 & 0.343 & 0.314 & 0.332 \\
					VILA \cite{ke2023vila}  & 0.652 & 0.663 & 0.548 & 0.561 & 0.413 & 0.443 \\
					UNIAA \cite{zhou2024uniaa}  & 0.472 & 0.390 & -& - & 0.261 & 0.260 \\
					Charm \cite{behrad2025charm} & 0.651 & 0.661 & 0.510 & 0.528 & 0.411 & 0.441 \\
					\midrule
					\midrule
					\textbf{FGAesQ} & \textbf{0.653} & \textbf{0.674} & \textbf{0.562} & \textbf{0.571} & \textbf{0.423} & \textbf{0.455} \\
					\bottomrule
			\end{tabular}}
\label{Tab4}
\end{table}

\noindent \textbf{Cross-dataset Evaluation.}
We further conduct generalization evaluations on three IAA benchmarks, including ICAA17K \cite{he2023thinking} (color aesthetics), AADB \cite{kong2016photo} (aesthetic attributes), and TAD66K \cite{he2022rethinking} (theme-specific aesthetics), as summarized in \cref{Tab4}. FGAesQ demonstrates superior generalization across all datasets, particularly exhibiting significant advantages on AADB. This suggests that training with fine-grained aesthetic comparisons enhances the perception of aesthetic attributes, enabling more accurate evaluation.

\begin{table}[t]
	\centering
	\caption{Ablation study on training strategies and model components. `w/o' is removal of the specified training or component. Metrics are (\textit{Acc}, \textit{F1})/2 for Pair and (\textit{s-Acc}, \textit{s-SRCC})/2 for Series.}
	\resizebox{0.81\linewidth}{!}{
		\begin{tabular}{l|cc|cc}
			\toprule
			\multirow{2}{*}{\textbf{Methods}} & \multicolumn{2}{c|}{\textbf{Coarse-grained}} & \multicolumn{2}{c}{\textbf{Fine-grained}} \\
			\cline{2-5}
			\rule{0pt}{11pt}
			& SRCC & PLCC & Pair & Series \\
			\midrule
			\midrule
			w/o Fine & 0.713& 0.726& 0.578& 0.364\\
			w/o Coarse & 0.031& 0.050& 0.565& 0.299\\
			Coarse \textbar{} Fine & 0.200 & 0.214 & 0.637 & 0.380 \\
			\midrule
			w/o DiffToken & 0.751& 0.760& 0.666& 0.423\\
			w/o CTAlign & 0.770 & 0.780 & 0.747 & 0.581 \\
			w/o RankReg & 0.769 & 0.781 & 0.742 & 0.571 \\
			\midrule
			\midrule
			\textbf{FGAesQ} & \textbf{0.770} & \textbf{0.781} & \textbf{0.753} & \textbf{0.600} \\
			\bottomrule
	\end{tabular}}
	\label{Tab5}
\end{table}

\begin{figure}[t]
	\centering
	\includegraphics[width=0.42\textwidth]{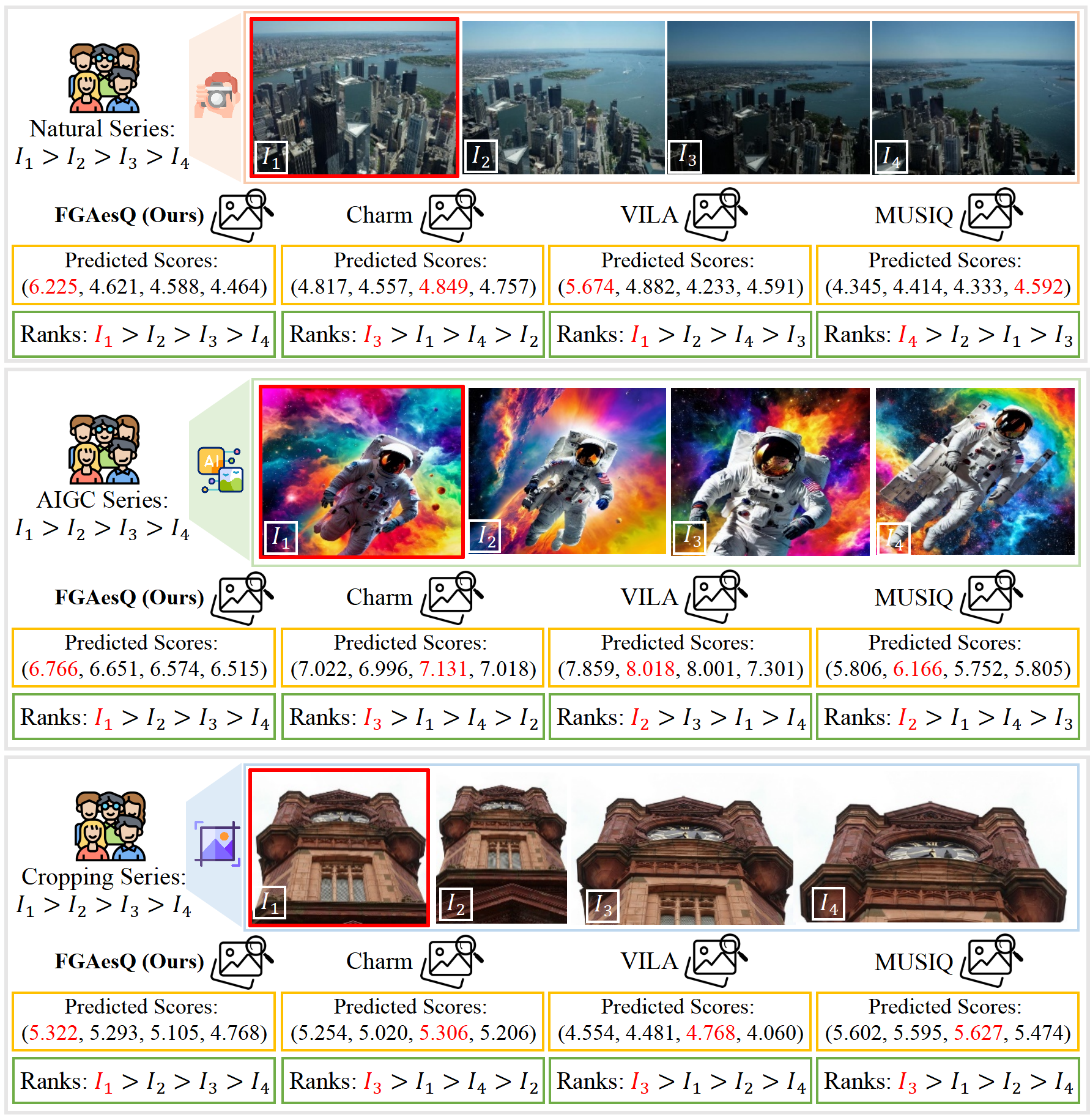}
	\caption{Visualization of evaluation results on three test series. Images arranged left-to-right in decreasing aesthetic quality. Red boxes and text indicate the best aesthetics.}
	\label{Fig5}
\end{figure}

\subsection{Ablation Study}
We conduct extensive ablations to validate training strategy and model components of FGAesQ, as listed in \cref{Tab5}.

\noindent \textbf{Training Strategy.}
We first evaluate FGAesQ with isolated training (w/o Fine and w/o Coarse). Performance drops when the corresponding training is absent, confirming the distinct nature of fine- and coarse-grained aesthetic data. The sequential training (Coarse \textbar{} Fine) also shows substantial degradation, demonstrating the effectiveness of two-stage training strategy for FGAesQ.

\noindent \textbf{Model Components.}
We further evaluate component contributions through model variants (w/o DiffToken, w/o CTAlign and w/o RankReg). Decreases in both pair-level and series-level performance demonstrate their effectiveness for fine-grained aesthetic modeling. The performance drops when removing text particularly validate the value of MLLM-generated comparative descriptions. Note that w/o RankReg refers to direct ranking training.

\subsection{Visual Analysis}
In \cref{Fig5}, we show the visualization results of FGAesQ, Charm \cite{behrad2025charm}, VILA \cite{ke2023vila}, and MUSIQ \cite{ke2021musiq} on three test series. FGAesQ not only accurately identifies the most aesthetic image (highlighted by red boxes), but also predicts rankings with better consistency with human annotations across the entire series. This demonstrates its superior capability in capturing fine-grained aesthetic nuances.

%% file: sec/6_conclusion.tex
\section{Conclusion}
In this paper, we have introduced the task of fine-grained image aesthetic assessment that discriminates subtle aesthetic differences among visually similar photo series. We contribute FGAesthetics, a large-scale dataset containing diverse image series with aesthetic rankings. Experiments on FGAesthetics reveal that existing IAA methods, while achieving excellent coarse-grained evaluation, fundamentally struggle with subtle aesthetic discrimination in fine-grained scenarios. Given this limitation, we further propose FGAesQ that learns discriminative scores from relative ranks, enabling superior evaluation performance in both fine- and coarse-grained scenarios. While encouraging results have been achieved in this work, fine-grained IAA remains challenging. We hope our contributions can inspire the community to rethink the problem with more technical approaches and broader perspectives.

%% file: sec/X_suppl.tex
\clearpage
\setcounter{page}{1}
\maketitlesupplementary

\section{FGAesthetics}

In this section, we provide supplementary details on the construction of FGAesthetics, including: (1) detailed source distribution statistics, (2) implementation procedures of series refinement and rank calibration, and (3) prompt engineering methodology for comparative text generation.

\subsection{Source Distribution of Data Collection}
\cref{sup-table1} presents the detailed source distribution of FGAesthetics. We collect 106,632 images from eight diverse sources spanning three distinct categories: natural images \cite{chang2016automatic,ying2021patch} , AI-generated content (AIGC) \cite{kirstain2023pick,zhang2025q,midjourney,fu2023dreamsim}, and cropping-based datasets \cite{wei2018good,zeng2019reliable}.  Through rigorous series refinement and rank calibration, the final dataset contains 32,217 images organized into 10,028 series, yielding an average of 4.47 pairs per series for fine-grained aesthetic discrimination. This diverse composition ensures that FGAesthetics encompasses comprehensive fine-grained aesthetic dimensions: natural photography quality, AI generation aesthetics, and compositional variations.

\begin{table}[h]
	\centering
	\caption{Source Distribution and Statistics of FGAesthetics}
	\resizebox{\linewidth}{!}{
		\begin{tabular}{l|ccccccc}
			\toprule
			\textbf{Sources} & \textbf{\makecell{Collected\\Img Count}} & \textbf{\makecell{After Series \\ Refinement}} & \textbf{\makecell{After Rank \\ Calibration}} & \textbf{\makecell{Series\\Count}} & \textbf{\makecell{Pair\\Count}} & \textbf{\makecell{Avg. Pairs\\per Series}} \\
			\midrule\midrule
			SPS \cite{chang2016automatic}         & 15,543 & 12,522 & 12,522 & 4,755 & 12,558  & 2.64 \\
			LSVQ \cite{ying2021patch}        &  2630 &   565  &    556 &   150 &    636  & 4.24 \\
			Pick-a-pic \cite{kirstain2023pick}  & 13,917 &  3,751 &  3,719 & 1,685 &  2,115  & 1.26 \\
			Q-Eval-100K \cite{zhang2025q} & 11,430 &   418  &    372 &   180 &    196  & 1.09 \\
			Midjourney \cite{midjourney}  & 11,352 &  2,815 &  2,599 &   909 &  2,124  & 2.34 \\
			NIGHTS \cite{fu2023dreamsim}     &  9,822 &  1,944 &  1,895 &   647 &  1,590  & 2.46 \\
			CPC \cite{wei2018good}         & 29,348 &  8,485 &  8,480 & 1,310 & 21,242  & 16.22 \\
			GAIC \cite{zeng2019reliable}      & 12,590 &  2,088 &  2,074 &   392 &  4,402 & 11.23 \\
			\midrule\midrule
			\textbf{Total} & \textbf{106,632} & \textbf{32,588} & \textbf{32,217} & \textbf{10,028} & \textbf{44,863} & \textbf{4.47} \\
			\bottomrule
		\end{tabular}
	}
	\label{sup-table1}
\end{table}

\subsection{Details of Series Refinement}
\begin{figure}[t]
	\centering
	\includegraphics[width=0.48\textwidth]{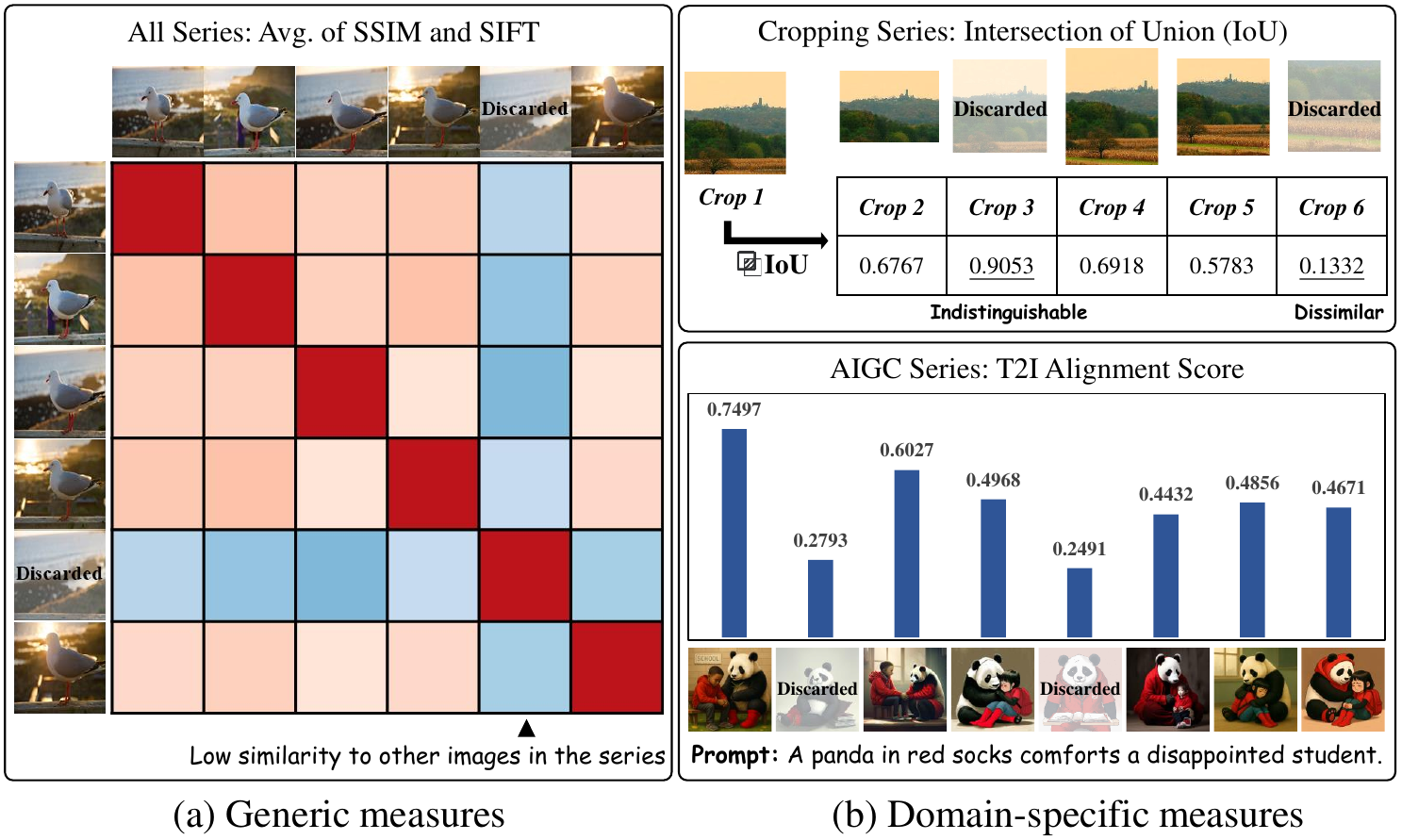}
	\caption{Detailed Procedure of Metrics Filtering. (a) Generic measures (average of SSIM and SIFT) exclude outliers within visually similar photo series. (b) Domain-specific measures (IoU for cropping series, T2I alignment scores for AIGC series) effectively identify nearly identical and dissimilar images.}
	\label{sup-fig1}
\end{figure}

\textbf{Metrics Filtering.} \cref{sup-fig1} illustrates the detailed procedure of Metrics Filtering. We employ both generic and domain-specific measures to screen the data. First, generic measures compute the average SSIM and SIFT scores between each image and all others within its series. We then rank all images across the entire dataset and filter out the bottom 30\% to exclude obvious outliers. For cropping series, we further calculate Intersection over Union (IoU) between crop frames and remove images with IoU $>0.8$ (indistinguishable) or IoU $<0.2$ (dissimilar). For AIGC series, we leverage original human-annotated T2I alignment scores to filter images severely misaligned with text prompts.

\noindent \textbf{MLLMs Checking using Gemini.}
Given the robust contextual understanding capabilities of MLLMs, we also employ Gemini-2.5-pro \cite{team2023gemini} to validate the filtered series. This is achieved by presenting all images within a series to the model, guided by a carefully crafted prompt that requires a structured output for systematic parsing:

\noindent \textit{\# system: \small You are a rigorous image series filtering expert. Your task is to filter a given series of images and identify a subset that adheres to the specified criteria.}

\noindent \textit{\# user: \small Your evaluation process is governed by the following criteria. Intra-group Similarity: Within the provided image series, identify the largest subset of images that are similar in theme, style, composition, and subject matter. They should appear as close variations of the same concept. If an image's theme significantly deviates from the other images, it must not be included in the final similar group. Your output must be a formatted LIST: ``selected\_uids": [``\textless uid\_a\textgreater", ``\textless uid\_b\textgreater", ...].}
 
\begin{figure}[t]
	\centering
	\includegraphics[width=0.48\textwidth]{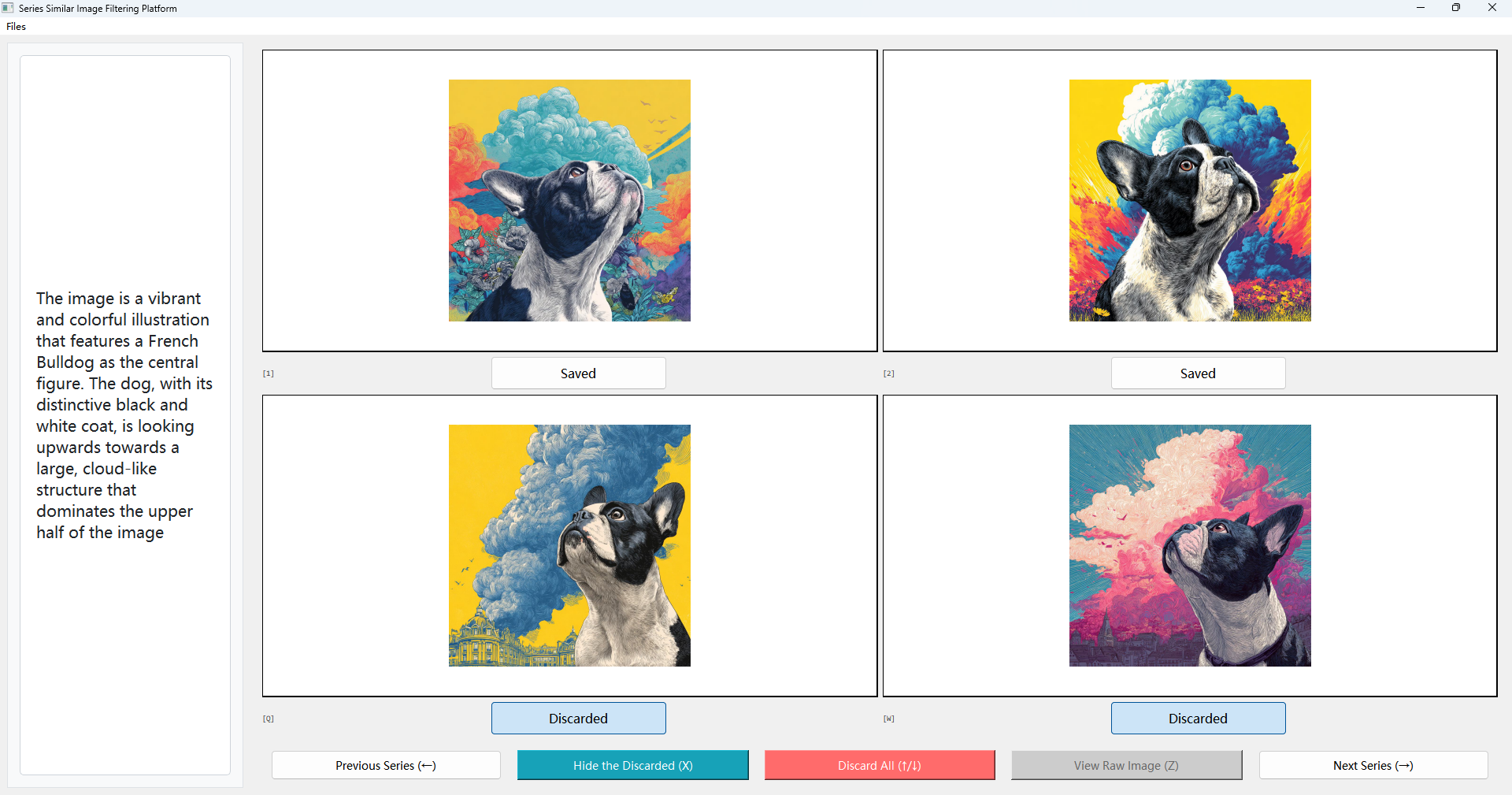}
	\caption{Annotation platform for Human Qualification. The platform displays all images from a single series on one page, enabling annotators to identify and filter outliers.}
	\label{sup-fig2}
\end{figure}

\noindent \textbf{Human Qualification.}
Following the automated filtering stages, comprehensive human qualification is implemented to ensure final data quality. All series undergo manual review by five human annotators via the annotation platform shown in \cref{sup-fig2}. Samples that fail to meet fine-grained aesthetic criteria are discarded, ensuring that only high-quality series proceed to the aesthetic ranking annotation stage.

\begin{figure}[t]
	\centering
	\includegraphics[width=0.48\textwidth]{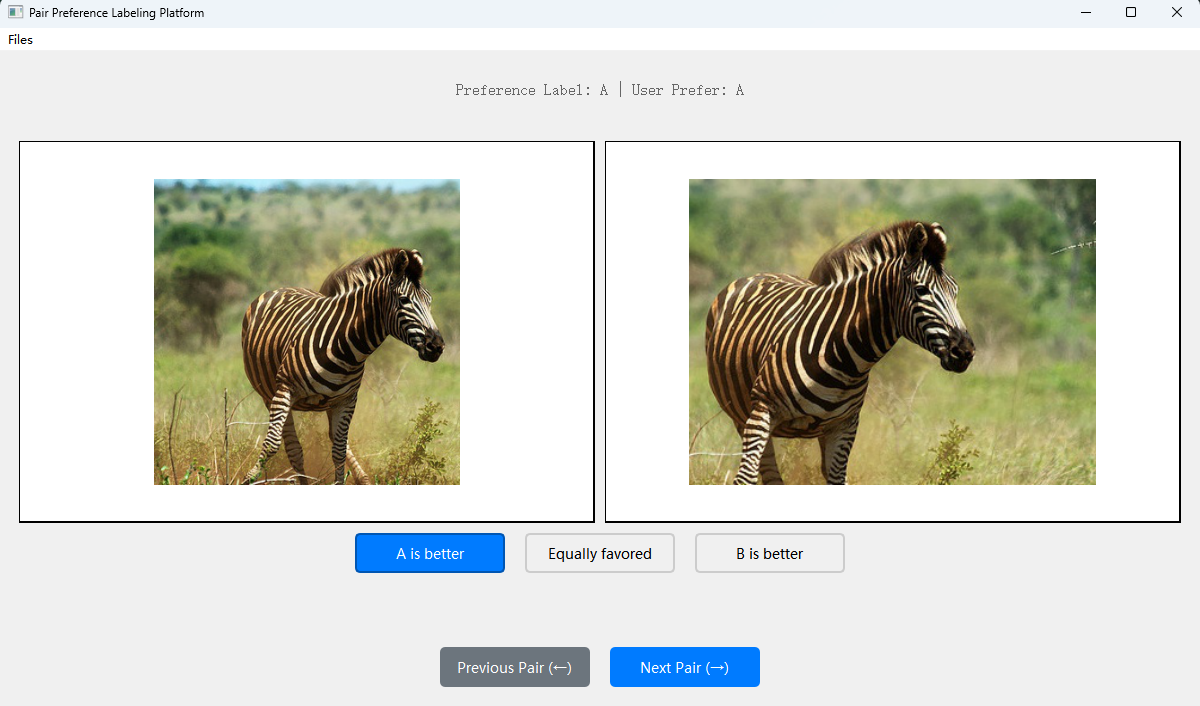}
	\caption{Annotation platform for Pairwise Comparison. The platform displays two images side-by-side for aesthetic comparison. Annotators make a forced-choice selection from three options: 'A is better', 'B is better', or 'Equally favored'.}
	\label{sup-fig3}
\end{figure}

\subsection{Details of Rank Calibration}
Based on the refined data, we perform pairwise comparisons within each series to obtain global aesthetic rankings. Specifically, for a series of length $n$, a total of $C_{n}^{2}$ image pairs are generated. Each pair is evaluated by 10 trained annotators for aesthetic comparison, determining which image is superior or indistinguishable, with the annotation platform shown in \cref{sup-fig3}. 

A total of 15 human evaluators (12 male, 3 female), all holding graduate degrees in computer science and actively engaged in IAA research, participated in the comprehensive annotation process for FGAesthetics. To mitigate potential biases from content variation, the annotation workload was systematically divided into three distinct sessions, each dedicated to one of the primary source categories: Natural, AIGC, and Cropping. The entire human annotation phase, which includes both the series refinement described before and the pairwise aesthetic comparisons, spanned 50 days.

\begin{figure}[t]
	\centering
	\includegraphics[width=0.49\textwidth]{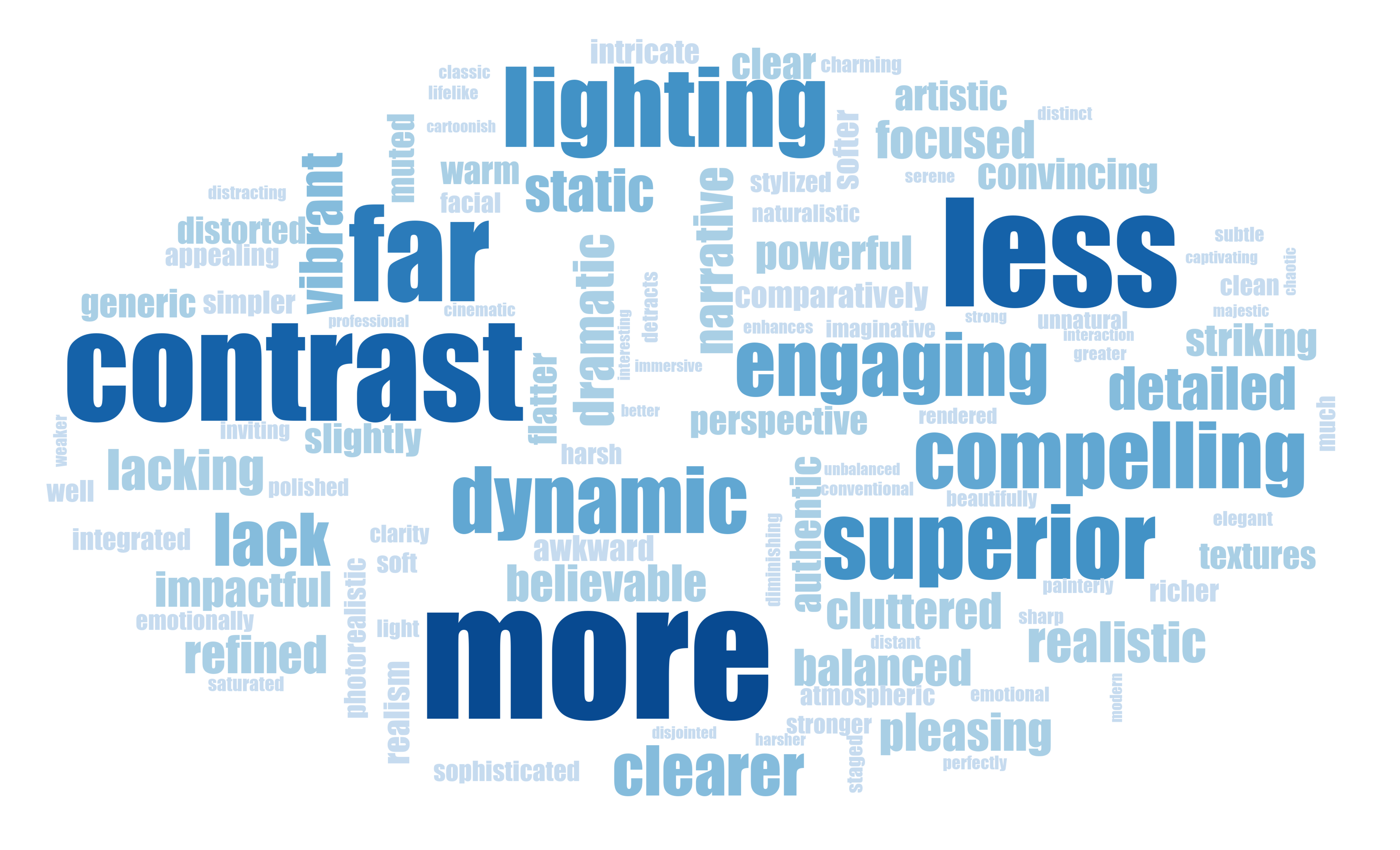}
	\caption{Word clouds visualizing most frequent rationales for MLLM-generated comparative textual descriptions.}
	\label{sup-fig4}
\end{figure}

\begin{table*}[t]
	\centering
	\small
	\caption{Backbone ablation results on fine-grained and coarse-grained IAA tasks. For fine-grained evaluation on FGAesthetics, we report pair-level local discrimination using \textit{Acc} and \textit{F1}, and series-level ranking quality using \textit{s-Acc} and \textit{s-SRCC} across three source categories (Natural, AIGC, Cropping). \textit{Pair} and \textit{Series} represent category-averaged values of (\textit{Acc}+\textit{F1})/2 and (\textit{s-Acc}+\textit{s-SRCC})/2, respectively. For coarse-grained evaluation, we report SRCC and PLCC on AVA \cite{murray2012ava}.}
	\resizebox{\linewidth}{!}{
		\begin{tabular}{c|cp{0.9cm}<{\centering}|cc|cp{0.9cm}<{\centering}|cc|cp{0.9cm}<{\centering}|cc|cc|cc}
			\toprule
			\multirow{2}{*}{\textbf{Backbone}} & \multicolumn{4}{c|}{\textbf{Natural}} & \multicolumn{4}{c|}{\textbf{AIGC}} & \multicolumn{4}{c|}{\textbf{Cropping}} & \multicolumn{2} {c|}{\textbf{Fine-grained}} & \multicolumn{2}{c}{\textbf{Coarse-grained}} \\
			\cline{2-17}
			\rule{0pt}{12pt}
			& \textit{Acc} & \textit{F1} & \textit{s-Acc} & \textit{s-SRCC} &\textit{Acc} & \textit{F1} & \textit{s-Acc} & \textit{s-SRCC} & \textit{Acc} & \textit{F1} & \textit{s-Acc} & \textit{s-SRCC} & \textit{Pair} & \textit{Series} & SRCC & PLCC \\
			\midrule\midrule
			ViT-B/16 & \textbf{0.779} & \textbf{0.779} & \textbf{0.753} & \textbf{0.729} & \textbf{0.709} & \textbf{0.707} & \textbf{0.561} & \textbf{0.482} & \textbf{0.774} & \textbf{0.773} & 0.488 & \textbf{0.590} & \textbf{0.753} & \textbf{0.600} & \textbf{0.770} & \textbf{0.781} \\
			ViT-B/32 & 0.711 & 0.711 & 0.634 & 0.503 & 0.653 & 0.651 & 0.445 & 0.301 & 0.740 & 0.740 & \textbf{0.491} & 0.548 & 0.701 & 0.487 & 0.747 & 0.760  \\
			\bottomrule
	\end{tabular}}
	\label{sup-table2}
\end{table*}

\begin{table*}[t]
	\centering
	\caption{Out-of-Distribution generalization performance. OOD generalization is evaluated by training FGAesQ with one source category excluded and testing on fine-grained (all three categories) and coarse-grained (AVA \cite{murray2012ava}) benchmarks.}
	\resizebox{\linewidth}{!}{
		\begin{tabular}{c|cp{0.9cm}<{\centering}|p{0.95cm}<{\centering}c|cp{0.9cm}<{\centering}|p{0.95cm}<{\centering}c|cp{0.9cm}<{\centering}|p{0.95cm}<{\centering}c|cc|cc}
			\toprule
			\multirow{2}{*}{\textbf{Backbone}} & \multicolumn{4}{c|}{\textbf{Natural}} & \multicolumn{4}{c|}{\textbf{AIGC}} & \multicolumn{4}{c|}{\textbf{Cropping}} & \multicolumn{2} {c|}{\textbf{Fine-grained}} & \multicolumn{2}{c}{\textbf{Coarse-grained}} \\
			\cline{2-17}
			\rule{0pt}{12pt}
			& \textit{Acc} & \textit{F1} & \textit{s-Acc} & \textit{s-SRCC} &\textit{Acc} & \textit{F1} & \textit{s-Acc} & \textit{s-SRCC} & \textit{Acc} & \textit{F1} & \textit{s-Acc} & \textit{s-SRCC} & \textit{Pair} & \textit{Series} & SRCC & PLCC \\
			\midrule\midrule
			FGAesQ                & 0.779 & 0.779 & 0.753 & 0.729 & 0.709 & 0.707 & 0.561 & 0.482 & 0.774 & 0.773 & 0.488 & 0.590 & 0.753 & 0.600 & 0.770 & 0.781  \\
			w/o Natural  & \textbf{0.745} & \textbf{0.745} & \textbf{0.710} & \textbf{0.630} & 0.704 & 0.703 & 0.569 & 0.455 & 0.771 & 0.771 & 0.488 & 0.583 & 0.740 & 0.573 & 0.770 & 0.782  \\
			w/o AIGC     & 0.767 & 0.766 & 0.720 & 0.675 & \textbf{0.687} & \textbf{0.685} & \textbf{0.542} & \textbf{0.425} & 0.765 & 0.765 & 0.482 & 0.577 & 0.739 & 0.570 & 0.776 & 0.786 \\
			w/o Cropping & 0.765 & 0.764 & 0.727 & 0.678 & 0.699 & 0.698 & 0.552 & 0.438 & \textbf{0.755} & \textbf{0.755} & \textbf{0.462} & \textbf{0.558} & 0.739 & 0.569 & 0.772 & 0.783 \\
			\bottomrule
	\end{tabular}}
	\label{sup-table3}
\end{table*}

\subsection{Prompt Engineering for Reasoning Generation}
Comparative textual descriptions are utilized to guide the visual model toward fine-grained aesthetic differences, enabling the CTAlign component of FGAesQ. To obtain these descriptions while minimizing human annotation burden, we leverage the robust reasoning capabilities of MLLMs. The detailed prompt template is provided below. Note that for AIGC and Cropping series, \textit{Original Images} and \textit{Shared Text Prompts} serve as additional contextual information.

\noindent \textit{\# system: \small You are an expert in digital art and aesthetics. You will be provided with a [Original Image, Shared Text Prompt] for context, two images, [Image A] and [Image B], and an aesthetic preference probability for Image A, [prob\_A\_preferred].} 
	
\noindent \textit{\# user: \small Your evaluation should be based on the concepts in the [Original Image, Shared Text Prompt], assessing creative execution, aesthetics, and other aspects of various aesthetic attributes. Your task is to generate a single string where two sentences are separated by ``\textbar \textbar". The first sentence describes Image A, and the second describes Image B, creating an implicit but powerful contrast.} 
	
\noindent \textit{\small You must follow these crucial rules: 
	\begin{enumerate}
		\item The sentences must not contain ``Image A," ``Image B," or similar direct labels; the comparison must be achieved through contrasting descriptions. 
		\item Each sentence must be concise (under 30 words) and use strong, comparative language (e.g., ``far more refined," ``lacks the depth," ``pales in comparison"). 
		\item Wording must not refer to the creation process (e.g., ``prompt," ``crop"); focus only on comparing the final images. 
	\end{enumerate}
}
	
\noindent \textit{\small Your output must be a formatted JSON object and must not contain any explanatory text outside of the JSON format. The JSON structure is as follows:}

{\small \color{gray}\texttt{\{\\
		\hspace*{3em}"description": <Your description>,\\
		\hspace*{3em}"prob\_A\_preferred": <the preference probability for Image A>\\
		\hspace*{1.3em}\}}}

\noindent \textit{\small For example, if the preference probability for Image A is 0.167 (B is preferred), your output should be:}

{\small \color{gray}\texttt{\{\\
		\hspace*{3em}"description": "This composition feels somewhat loose and the subject more distant, weakening the overall emotional impact.||By contrast, this one uses a much tighter and more intimate focus, creating a significantly more compelling narrative.",\\
		\hspace*{3em}"prob\_A\_preferred": 0.167\\
		\hspace*{1.3em}\}}}

\noindent \textit{\small Notably, the Comparison Direction of these two descriptions must be consistent with prob\_A\_preferred, that is, there cannot be a situation like prob\_A\_preferred is less than 0.5 but description A is more positive than description B.}

To validate the quality of generated descriptions, we conduct a human evaluation. Three independent annotators assess 100 randomly sampled descriptions for plausibility, achieving an agreement rate of 93\%. This confirms that our prompt engineering approach effectively generates meaningful and accurate explanations capturing fine-grained aesthetic differences. In \cref{sup-fig4}, we further visualize the word cloud of MLLM-generated descriptions, where explicit comparative terms (e.g., ``more", ``superior", ``less") dominate the vocabulary.

\section{FGAesQ}
Following the implementation details and extensive experiments described in the paper, we present supplementary ablation studies to further validate the design choices of FGAesQ. Specifically, we investigate the impact of backbone architecture, Out-of-Distribution (OOD) generalization capability, and different DiffToken configurations.
\subsection{Ablation of Backbone}
We first investigate the impact of backbone architecture by replacing ViT-B/16 with ViT-B/32, a variant with a larger patch size. As shown in \cref{sup-table2}, ViT-B/16-based FGAesQ consistently outperforms ViT-B/32 across most fine-grained and coarse-grained evaluations. The performance gap can be attributed to the smaller patch size (16$\times$16 vs. 32$\times$32) of ViT-B/16, which enables the model to capture more granular visual details. This is crucial for discerning subtle aesthetic differences required for both fine-grained ranking and coarse-grained scoring.

\subsection{Out-of-Distribution (OOD) Generalization}
We further conduct an Out-of-Distribution (OOD) generalization analysis of FGAesQ, as illustrated in \cref{sup-table3}. Specifically, we evaluate FGAesQ trained with one source category excluded and test its performance on both fine-grained and coarse-grained IAA tasks. The results reveal that performance degradation is most pronounced on the test set corresponding to the excluded training category. For instance, when trained without Natural data, the model shows the largest performance drop on Natural evaluation (from 0.779/0.753 to 0.745/0.710 for Acc/s-Acc), while maintaining relatively stable performance on AIGC and Cropping. This pattern holds consistently across all three categories, demonstrating the distinct nature of each data source. Coarse-grained performance remains largely stable across all training configurations, with SRCC and PLCC consistently above 0.77 and 0.78, respectively. Notably, excluding AIGC data even leads to a slight improvement in coarse-grained metrics, which can be attributed to the fact that AVA predominantly consists of natural images.

\subsection{Ablation of DiffToken}
\begin{figure}[t]
	\centering
	\includegraphics[width=0.48\textwidth]{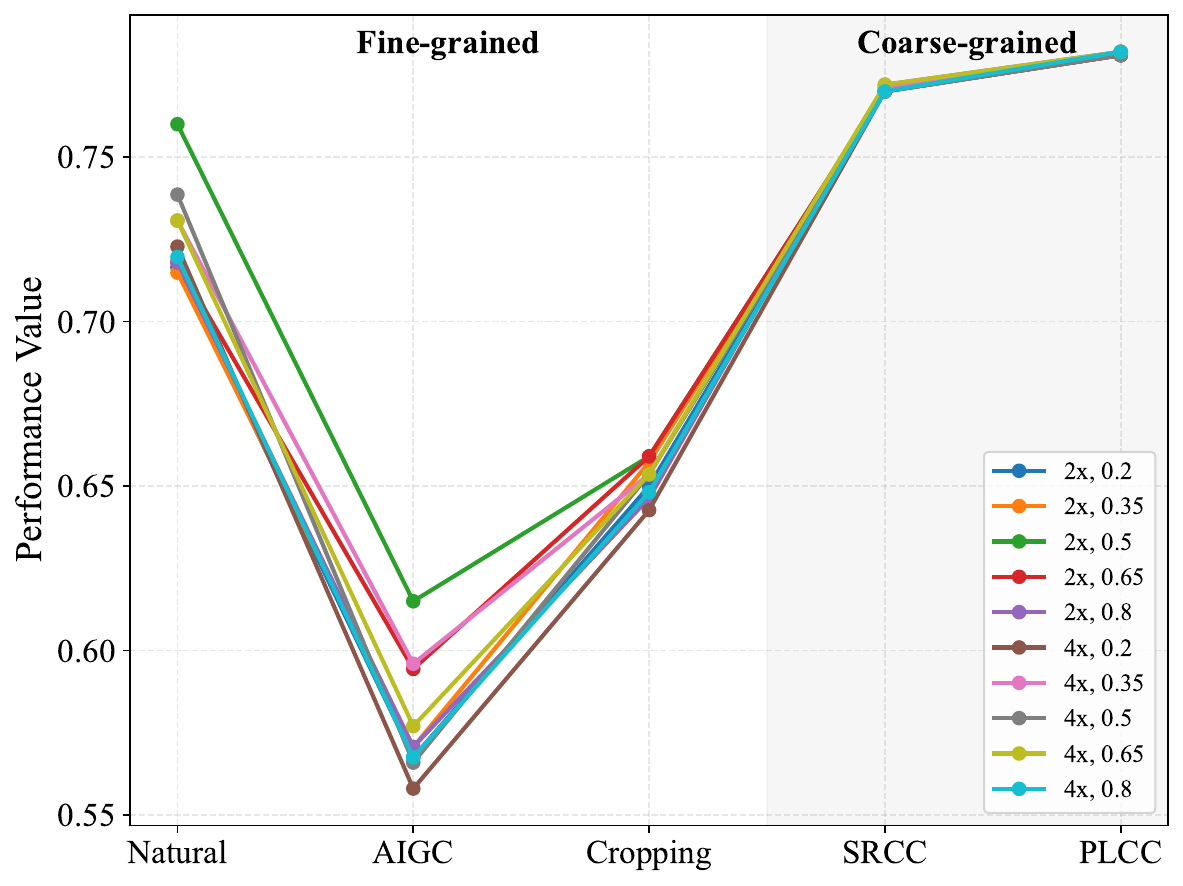}
	\caption{Ablation study on DiffToken configuration. Performance is evaluated across different difference localization patch sizes and percentile thresholds for identifying aesthetics-decisive regions. Fine-grained metrics show pair- and series-averaged performance across Natural, AIGC, and Cropping categories. Coarse-grained metrics report SRCC and PLCC on AVA \cite{murray2012ava}.}
	\label{sup-fig5}
\end{figure}

As introduced in the main paper, the Difference-preserved Tokenization (DiffToken) module is governed by two key hyperparameters: the difference localization patch size and the percentile parameter $p$, which determines the proportion of aesthetics-decisive regions. We conduct an ablation study evaluating two patch sizes, 32$\times$32 (2$\times$ ViT patch size) and 64$\times$64 (4$\times$), combined with five percentile values: $p \in \{0.2, 0.35, 0.5, 0.65, 0.8\}$, as illustrated in \cref{sup-fig5}.

The results yield several key insights. First, the optimal configuration is 32$\times$32 with $p=0.5$, achieving the best overall performance across all tasks. Second, performance exhibits a clear inverted-U pattern with respect to $p$, peaking around 0.5 and degrading at extremes. A small $p$ (e.g., 0.2) fails to capture sufficient aesthetically-decisive regions, while too large $p$ (e.g., 0.8) forces aggressive downsampling of non-decisive regions to meet the token constraints, losing critical global compositional information. Third, the finer 32$\times$32 localization consistently outperforms 64$\times$64, as smaller patches more effectively detect subtle local variations crucial for fine-grained IAA. Finally, coarse-grained performance (SRCC/PLCC on AVA) remains stable across all configurations, confirming that DiffToken's hyperparameters primarily affect fine-grained discrimination while having negligible impact on absolute aesthetic assessment.

\begin{figure*}[b]
	\centering
	\includegraphics[width=\textwidth]{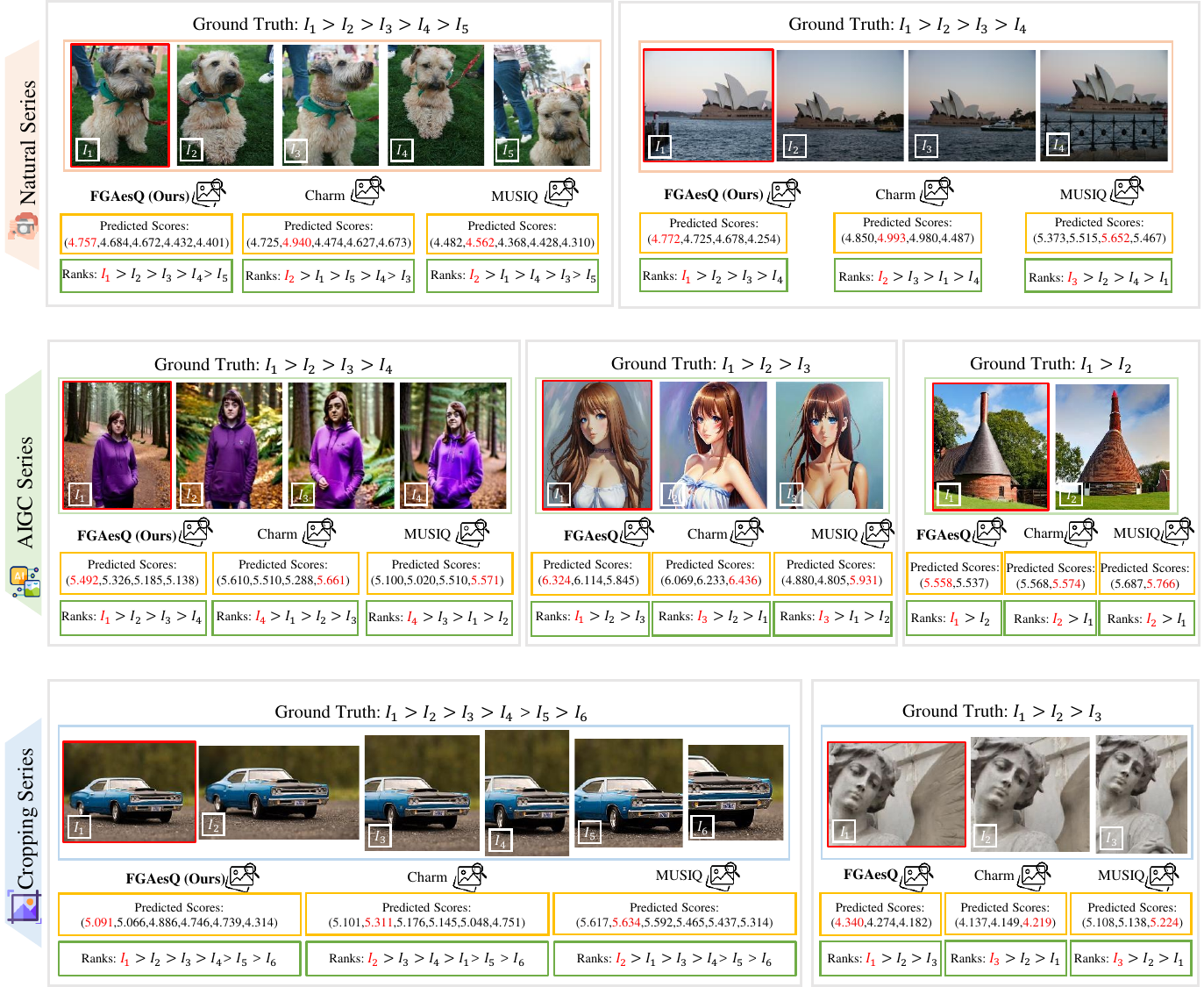}
	\caption{Additional visualization examples of FGAesQ on test series from Natural, AIGC, and Cropping categories.}
	\label{sup-fig6}
\end{figure*}

\section{Additional Visual Examples}
In \cref{sup-fig6}, we present additional visualization examples demonstrating FGAesQ's fine-grained aesthetic discrimination capabilities. The visualizations showcase test series from Natural, AIGC, and Cropping categories, where FGAesQ consistently produces accurate rankings that closely align with human aesthetic judgments. Compared to state-of-art IAA methods (Charm \cite{behrad2025charm}, MUSIQ \cite{ke2021musiq}), our approach exhibits superior sensitivity to subtle aesthetic variations, including lighting conditions, color harmony, and compositional balance. This robust performance across different evaluation scenarios highlights FGAesQ's versatility and practical value for real-world applications, such as photo album management and curation, text-to-image generation optimization, and automated composition refinement for photography enhancement.

\section{Limitations}
\textbf{Dependency on Human Annotations.} The reliance on extensive human annotations for fine-grained aesthetic discrimination poses scalability challenges and introduces potential subjective biases. Developing automated or semi-automated annotation strategies that maintain annotation quality while reducing human effort remains an important direction for future work.

\noindent \textbf{Interpretability and Feedbacks.} While FGAesQ achieves accurate fine-grained aesthetic discrimination, generating specific and actionable feedback to explain its predictions remains challenging. Developing methods that articulate precise aesthetic rationales, such as identifying compositional flaws or suggesting targeted improvements, would significantly enhance the practical applicability of fine-grained IAA models. This promising yet challenging direction merits further exploration.

%% file: main.bbl
\begin{thebibliography}{53}
\providecommand{\natexlab}[1]{#1}
\providecommand{\url}[1]{\texttt{#1}}
\expandafter\ifx\csname urlstyle\endcsname\relax
  \providecommand{\doi}[1]{doi: #1}\else
  \providecommand{\doi}{doi: \begingroup \urlstyle{rm}\Url}\fi

\bibitem[Achiam et~al.(2023)Achiam, Adler, Agarwal, Ahmad, Akkaya, Aleman,
  Almeida, Altenschmidt, Altman, Anadkat, et~al.]{achiam2023gpt}
Josh Achiam, Steven Adler, Sandhini Agarwal, Lama Ahmad, Ilge Akkaya,
  Florencia~Leoni Aleman, Diogo Almeida, Janko Altenschmidt, Sam Altman,
  Shyamal Anadkat, et~al.
\newblock Gpt-4 technical report.
\newblock \emph{arXiv preprint arXiv:2303.08774}, 2023.

\bibitem[Ba et~al.(2025)Ba, Zhang, Bai, Mo, Liang, Su, and
  Wen]{ba2025enhancing}
Ying Ba, Tianyu Zhang, Yalong Bai, Wenyi Mo, Tao Liang, Bing Su, and Ji-Rong
  Wen.
\newblock Enhancing reward models for high-quality image generation: Beyond
  text-image alignment.
\newblock In \emph{Proceedings of the IEEE/CVF International Conference on
  Computer Vision}, pages 19022--19031, 2025.

\bibitem[Behrad et~al.(2025)Behrad, Tuytelaars, and Wagemans]{behrad2025charm}
Fatemeh Behrad, Tinne Tuytelaars, and Johan Wagemans.
\newblock Charm: The missing piece in vit fine-tuning for image aesthetic
  assessment.
\newblock In \emph{Proceedings of the Computer Vision and Pattern Recognition
  Conference}, pages 7815--7824, 2025.

\bibitem[Bradley and Terry(1952)]{bradley1952rank}
Ralph~Allan Bradley and Milton~E Terry.
\newblock Rank analysis of incomplete block designs: I. the method of paired
  comparisons.
\newblock \emph{Biometrika}, 39\penalty0 (3/4):\penalty0 324--345, 1952.

\bibitem[Chang et~al.(2016)Chang, Yu, Wang, Ashley, and
  Finkelstein]{chang2016automatic}
Huiwen Chang, Fisher Yu, Jue Wang, Douglas Ashley, and Adam Finkelstein.
\newblock Automatic triage for a photo series.
\newblock \emph{ACM Transactions on Graphics (TOG)}, 35\penalty0 (4):\penalty0
  1--10, 2016.

\bibitem[Datta et~al.(2006)Datta, Joshi, Li, and Wang]{datta2006studying}
Ritendra Datta, Dhiraj Joshi, Jia Li, and James~Z Wang.
\newblock Studying aesthetics in photographic images using a computational
  approach.
\newblock In \emph{European conference on computer vision}, pages 288--301.
  Springer, 2006.

\bibitem[Datta et~al.(2008)Datta, Li, and Wang]{datta2008algorithmic}
Ritendra Datta, Jia Li, and James~Z Wang.
\newblock Algorithmic inferencing of aesthetics and emotion in natural images:
  An exposition.
\newblock In \emph{2008 15th IEEE international conference on image
  processing}, pages 105--108. IEEE, 2008.

\bibitem[Deng et~al.(2017)Deng, Loy, and Tang]{deng2017image}
Yubin Deng, Chen~Change Loy, and Xiaoou Tang.
\newblock Image aesthetic assessment: An experimental survey.
\newblock \emph{IEEE Signal Processing Magazine}, 34\penalty0 (4):\penalty0
  80--106, 2017.

\bibitem[Fu et~al.(2023)Fu, Tamir, Sundaram, Chai, Zhang, Dekel, and
  Isola]{fu2023dreamsim}
Stephanie Fu, Netanel~Y Tamir, Shobhita Sundaram, Lucy Chai, Richard Zhang,
  Tali Dekel, and Phillip Isola.
\newblock Dreamsim: learning new dimensions of human visual similarity using
  synthetic data.
\newblock In \emph{Proceedings of the 37th International Conference on Neural
  Information Processing Systems}, pages 50742--50768, 2023.

\bibitem[Havtorn et~al.(2023)Havtorn, Royer, Blankevoort, and
  Bejnordi]{havtorn2023msvit}
Jakob~Drachmann Havtorn, Am{\'e}lie Royer, Tijmen Blankevoort, and
  Babak~Ehteshami Bejnordi.
\newblock Msvit: Dynamic mixed-scale tokenization for vision transformers.
\newblock In \emph{Proceedings of the IEEE/CVF International Conference on
  Computer Vision}, pages 838--848, 2023.

\bibitem[He et~al.(2022)He, Zhang, Xie, Jiang, and Ming]{he2022rethinking}
Shuai He, Yongchang Zhang, Rui Xie, Dongxiang Jiang, and Anlong Ming.
\newblock Rethinking image aesthetics assessment: Models, datasets and
  benchmarks.
\newblock In \emph{IJCAI}, pages 942--948, 2022.

\bibitem[He et~al.(2023)He, Ming, Li, Sun, Zheng, and Ma]{he2023thinking}
Shuai He, Anlong Ming, Yaqi Li, Jinyuan Sun, ShunTian Zheng, and Huadong Ma.
\newblock Thinking image color aesthetics assessment: Models, datasets and
  benchmarks.
\newblock In \emph{Proceedings of the IEEE/CVF International Conference on
  Computer Vision}, pages 21838--21847, 2023.

\bibitem[Hessel et~al.(2021)Hessel, Holtzman, Forbes, Le~Bras, and
  Choi]{hessel2021clipscore}
Jack Hessel, Ari Holtzman, Maxwell Forbes, Ronan Le~Bras, and Yejin Choi.
\newblock Clipscore: A reference-free evaluation metric for image captioning.
\newblock In \emph{Proceedings of the 2021 Conference on Empirical Methods in
  Natural Language Processing}, pages 7514--7528, 2021.

\bibitem[Holz(2023)]{midjourney}
David Holz.
\newblock Midjourney.
\newblock \url{https://www.midjourney}, 2023.

\bibitem[Hosu et~al.(2019)Hosu, Goldlucke, and Saupe]{hosu2019effective}
Vlad Hosu, Bastian Goldlucke, and Dietmar Saupe.
\newblock Effective aesthetics prediction with multi-level spatially pooled
  features.
\newblock In \emph{proceedings of the IEEE/CVF conference on computer vision
  and pattern recognition}, pages 9375--9383, 2019.

\bibitem[Huang et~al.(2024)Huang, Yuan, Sheng, Yang, Wu, Chen, Yang, Li, and
  Lin]{huang2024aesbench}
Yipo Huang, Quan Yuan, Xiangfei Sheng, Zhichao Yang, Haoning Wu, Pengfei Chen,
  Yuzhe Yang, Leida Li, and Weisi Lin.
\newblock Aesbench: An expert benchmark for multimodal large language models on
  image aesthetics perception.
\newblock \emph{arXiv preprint arXiv:2401.08276}, 2024.

\bibitem[Ke et~al.(2021)Ke, Wang, Wang, Milanfar, and Yang]{ke2021musiq}
Junjie Ke, Qifei Wang, Yilin Wang, Peyman Milanfar, and Feng Yang.
\newblock Musiq: Multi-scale image quality transformer.
\newblock In \emph{Proceedings of the IEEE/CVF international conference on
  computer vision}, pages 5148--5157, 2021.

\bibitem[Ke et~al.(2023)Ke, Ye, Yu, Wu, Milanfar, and Yang]{ke2023vila}
Junjie Ke, Keren Ye, Jiahui Yu, Yonghui Wu, Peyman Milanfar, and Feng Yang.
\newblock Vila: Learning image aesthetics from user comments with
  vision-language pretraining.
\newblock In \emph{Proceedings of the IEEE/CVF Conference on Computer Vision
  and Pattern Recognition}, pages 10041--10051, 2023.

\bibitem[Ke et~al.(2006)Ke, Tang, and Jing]{ke2006design}
Yan Ke, Xiaoou Tang, and Feng Jing.
\newblock The design of high-level features for photo quality assessment.
\newblock In \emph{2006 IEEE Computer Society Conference on Computer Vision and
  Pattern Recognition (CVPR'06)}, pages 419--426. IEEE, 2006.

\bibitem[Kirstain et~al.(2023)Kirstain, Polyak, Singer, Matiana, Penna, and
  Levy]{kirstain2023pick}
Yuval Kirstain, Adam Polyak, Uriel Singer, Shahbuland Matiana, Joe Penna, and
  Omer Levy.
\newblock Pick-a-pic: An open dataset of user preferences for text-to-image
  generation.
\newblock \emph{Advances in neural information processing systems},
  36:\penalty0 36652--36663, 2023.

\bibitem[Kong et~al.(2016)Kong, Shen, Lin, Mech, and Fowlkes]{kong2016photo}
Shu Kong, Xiaohui Shen, Zhe Lin, Radomir Mech, and Charless Fowlkes.
\newblock Photo aesthetics ranking network with attributes and content
  adaptation.
\newblock In \emph{European conference on computer vision}, pages 662--679.
  Springer, 2016.

\bibitem[Li et~al.(2025)Li, Wang, Sun, Bai, and Chu]{li2025next}
Mingxing Li, Rui Wang, Lei Sun, Yancheng Bai, and Xiangxiang Chu.
\newblock Next token is enough: Realistic image quality and aesthetic scoring
  with multimodal large language model.
\newblock \emph{arXiv preprint arXiv:2503.06141}, 2025.

\bibitem[Loui and Savakis(2003)]{loui2003automated}
Alexander~C Loui and Andreas Savakis.
\newblock Automated event clustering and quality screening of consumer pictures
  for digital albuming.
\newblock \emph{IEEE Transactions on Multimedia}, 5\penalty0 (3):\penalty0
  390--402, 2003.

\bibitem[Lu et~al.(2015)Lu, Lin, Shen, Mech, and Wang]{lu2015deep}
Xin Lu, Zhe Lin, Xiaohui Shen, Radomir Mech, and James~Z Wang.
\newblock Deep multi-patch aggregation network for image style, aesthetics, and
  quality estimation.
\newblock In \emph{Proceedings of the IEEE international conference on computer
  vision}, pages 990--998, 2015.

\bibitem[Luo et~al.(2011)Luo, Wang, and Tang]{luo2011content}
Wei Luo, Xiaogang Wang, and Xiaoou Tang.
\newblock Content-based photo quality assessment.
\newblock In \emph{2011 international conference on computer vision}, pages
  2206--2213. IEEE, 2011.

\bibitem[Murray et~al.(2012)Murray, Marchesotti, and Perronnin]{murray2012ava}
Naila Murray, Luca Marchesotti, and Florent Perronnin.
\newblock Ava: A large-scale database for aesthetic visual analysis.
\newblock In \emph{2012 IEEE conference on computer vision and pattern
  recognition}, pages 2408--2415. IEEE, 2012.

\bibitem[Radford et~al.(2021)Radford, Kim, Hallacy, Ramesh, Goh, Agarwal,
  Sastry, Askell, Mishkin, Clark, et~al.]{radford2021learning}
Alec Radford, Jong~Wook Kim, Chris Hallacy, Aditya Ramesh, Gabriel Goh,
  Sandhini Agarwal, Girish Sastry, Amanda Askell, Pamela Mishkin, Jack Clark,
  et~al.
\newblock Learning transferable visual models from natural language
  supervision.
\newblock In \emph{International conference on machine learning}, pages
  8748--8763. PmLR, 2021.

\bibitem[Ren et~al.(2020)Ren, Shen, Lin, and Mech]{ren2020best}
Jian Ren, Xiaohui Shen, Zhe Lin, and Radomir Mech.
\newblock Best frame selection in a short video.
\newblock In \emph{Proceedings of the IEEE/CVF Winter Conference on
  applications of computer vision}, pages 3212--3221, 2020.

\bibitem[Rezatofighi et~al.(2019)Rezatofighi, Tsoi, Gwak, Sadeghian, Reid, and
  Savarese]{rezatofighi2019generalized}
Hamid Rezatofighi, Nathan Tsoi, JunYoung Gwak, Amir Sadeghian, Ian Reid, and
  Silvio Savarese.
\newblock Generalized intersection over union: A metric and a loss for bounding
  box regression.
\newblock In \emph{Proceedings of the IEEE/CVF conference on computer vision
  and pattern recognition}, pages 658--666, 2019.

\bibitem[Ronen et~al.(2023)Ronen, Levy, and Golbert]{ronen2023vision}
Tomer Ronen, Omer Levy, and Avram Golbert.
\newblock Vision transformers with mixed-resolution tokenization.
\newblock In \emph{Proceedings of the IEEE/CVF Conference on Computer Vision
  and Pattern Recognition}, pages 4613--4622, 2023.

\bibitem[Sheng et~al.(2023)Sheng, Li, Chen, Wu, Dong, Yang, Xu, Li, and
  Shi]{sheng2023aesclip}
Xiangfei Sheng, Leida Li, Pengfei Chen, Jinjian Wu, Weisheng Dong, Yuzhe Yang,
  Liwu Xu, Yaqian Li, and Guangming Shi.
\newblock Aesclip: Multi-attribute contrastive learning for image aesthetics
  assessment.
\newblock In \emph{Proceedings of the 31st ACM International Conference on
  Multimedia}, pages 1117--1126, 2023.

\bibitem[Sheng et~al.(2025{\natexlab{a}})Sheng, Duan, Pan, Huang, Yang, Chen,
  and Li]{sheng2025tuningiqa}
Xiangfei Sheng, Zhichao Duan, Xiaofeng Pan, Yipo Huang, Zhichao Yang, Pengfei
  Chen, and Leida Li.
\newblock Tuningiqa: Fine-grained blind image quality assessment for
  livestreaming camera tuning.
\newblock \emph{arXiv preprint arXiv:2508.17965}, 2025{\natexlab{a}}.

\bibitem[Sheng et~al.(2025{\natexlab{b}})Sheng, Pan, Yang, Chen, and
  Li]{sheng2025fine}
Xiangfei Sheng, Xiaofeng Pan, Zhichao Yang, Pengfei Chen, and Leida Li.
\newblock Fine-grained image quality assessment for perceptual image
  restoration.
\newblock \emph{arXiv preprint arXiv:2508.14475}, 2025{\natexlab{b}}.

\bibitem[Sun et~al.(2017)Sun, Chao, Kuo, and Hsu]{sun2017photo}
Wei-Tse Sun, Ting-Hsuan Chao, Yin-Hsi Kuo, and Winston~H Hsu.
\newblock Photo filter recommendation by category-aware aesthetic learning.
\newblock \emph{IEEE Transactions on Multimedia}, 19\penalty0 (8):\penalty0
  1870--1880, 2017.

\bibitem[Tai et~al.(2024)Tai, Fan, Zhang, and Liu]{tai2024link}
Yan Tai, Weichen Fan, Zhao Zhang, and Ziwei Liu.
\newblock Link-context learning for multimodal llms.
\newblock In \emph{Proceedings of the IEEE/CVF Conference on Computer Vision
  and Pattern Recognition}, pages 27176--27185, 2024.

\bibitem[Talebi and Milanfar(2018)]{talebi2018nima}
Hossein Talebi and Peyman Milanfar.
\newblock Nima: Neural image assessment.
\newblock \emph{IEEE transactions on image processing}, 27\penalty0
  (8):\penalty0 3998--4011, 2018.

\bibitem[Team et~al.(2023)Team, Anil, Borgeaud, Alayrac, Yu, Soricut,
  Schalkwyk, Dai, Hauth, Millican, et~al.]{team2023gemini}
Gemini Team, Rohan Anil, Sebastian Borgeaud, Jean-Baptiste Alayrac, Jiahui Yu,
  Radu Soricut, Johan Schalkwyk, Andrew~M Dai, Anja Hauth, Katie Millican,
  et~al.
\newblock Gemini: a family of highly capable multimodal models.
\newblock \emph{arXiv preprint arXiv:2312.11805}, 2023.

\bibitem[Wang et~al.(2023)Wang, Chan, and Loy]{wang2023exploring}
Jianyi Wang, Kelvin~CK Chan, and Chen~Change Loy.
\newblock Exploring clip for assessing the look and feel of images.
\newblock In \emph{Proceedings of the AAAI conference on artificial
  intelligence}, pages 2555--2563, 2023.

\bibitem[Wang et~al.(2004)Wang, Bovik, Sheikh, and Simoncelli]{wang2004image}
Zhou Wang, Alan~C Bovik, Hamid~R Sheikh, and Eero~P Simoncelli.
\newblock Image quality assessment: from error visibility to structural
  similarity.
\newblock \emph{IEEE transactions on image processing}, 13\penalty0
  (4):\penalty0 600--612, 2004.

\bibitem[Wei et~al.(2018)Wei, Zhang, Shen, Lin, Mech, Hoai, and
  Samaras]{wei2018good}
Zijun Wei, Jianming Zhang, Xiaohui Shen, Zhe Lin, Radomir Mech, Minh Hoai, and
  Dimitris Samaras.
\newblock Good view hunting: Learning photo composition from dense view pairs.
\newblock In \emph{Proceedings of the IEEE conference on computer vision and
  pattern recognition}, pages 5437--5446, 2018.

\bibitem[Wu et~al.(2024{\natexlab{a}})Wu, Zhang, Zhang, Chen, Liao, Li, Gao,
  Wang, Zhang, Sun, et~al.]{wu2024q}
Haoning Wu, Zicheng Zhang, Weixia Zhang, Chaofeng Chen, Liang Liao, Chunyi Li,
  Yixuan Gao, Annan Wang, Erli Zhang, Wenxiu Sun, et~al.
\newblock Q-align: Teaching lmms for visual scoring via discrete text-defined
  levels.
\newblock In \emph{International Conference on Machine Learning}, pages
  54015--54029. PMLR, 2024{\natexlab{a}}.

\bibitem[Wu et~al.(2024{\natexlab{b}})Wu, Wang, Li, Zhang, and Xue]{wu2024goal}
Jiarui Wu, Yujin Wang, Lingen Li, Fan Zhang, and Tianfan Xue.
\newblock Goal conditioned reinforcement learning for photo finishing tuning.
\newblock \emph{Advances in Neural Information Processing Systems},
  37:\penalty0 46294--46318, 2024{\natexlab{b}}.

\bibitem[Xia et~al.(2008)Xia, Liu, Wang, Zhang, and Li]{xia2008listwise}
Fen Xia, Tie-Yan Liu, Jue Wang, Wensheng Zhang, and Hang Li.
\newblock Listwise approach to learning to rank: theory and algorithm.
\newblock In \emph{Proceedings of the 25th international conference on Machine
  learning}, pages 1192--1199, 2008.

\bibitem[Yang et~al.(2024{\natexlab{a}})Yang, Li, Chen, Wu, and
  Dong]{yang2024semantics}
Zhichao Yang, Leida Li, Pengfei Chen, Jinjian Wu, and Weisheng Dong.
\newblock Semantics-aware image aesthetics assessment using tag matching and
  contrastive ranking.
\newblock In \emph{Proceedings of the 32nd ACM International Conference on
  Multimedia}, pages 2632--2641, 2024{\natexlab{a}}.

\bibitem[Yang et~al.(2024{\natexlab{b}})Yang, Li, Yang, Li, and Lin]{10168279}
Zhichao Yang, Leida Li, Yuzhe Yang, Yaqian Li, and Weisi Lin.
\newblock Multi-level transitional contrast learning for personalized image
  aesthetics assessment.
\newblock \emph{IEEE Transactions on Multimedia}, 26:\penalty0 1944--1956,
  2024{\natexlab{b}}.

\bibitem[Yang et~al.(2025{\natexlab{a}})Yang, Gu, Wang, Lin, Sheng, Chen, and
  Li]{yang2025longt2ibench}
Zhichao Yang, Tianjiao Gu, Jianjie Wang, Feiyu Lin, Xiangfei Sheng, Pengfei
  Chen, and Leida Li.
\newblock Longt2ibench: A benchmark for evaluating long text-to-image
  generation with graph-structured annotations.
\newblock \emph{arXiv preprint arXiv:2512.09271}, 2025{\natexlab{a}}.

\bibitem[Yang et~al.(2025{\natexlab{b}})Yang, Li, Chen, Wu, and
  Valenzise]{yang2025language}
Zhichao Yang, Leida Li, Pengfei Chen, Jinjian Wu, and Giuseppe Valenzise.
\newblock Language-guided visual perception disentanglement for image quality
  assessment and conditional image generation.
\newblock \emph{arXiv preprint arXiv:2503.02206}, 2025{\natexlab{b}}.

\bibitem[Yi et~al.(2023)Yi, Tian, Gu, Lai, and Rosin]{yi2023towards}
Ran Yi, Haoyuan Tian, Zhihao Gu, Yu-Kun Lai, and Paul~L Rosin.
\newblock Towards artistic image aesthetics assessment: a large-scale dataset
  and a new method.
\newblock In \emph{Proceedings of the IEEE/CVF Conference on Computer Vision
  and Pattern Recognition}, pages 22388--22397, 2023.

\bibitem[Ying et~al.(2021)Ying, Mandal, Ghadiyaram, and Bovik]{ying2021patch}
Zhenqiang Ying, Maniratnam Mandal, Deepti Ghadiyaram, and Alan Bovik.
\newblock Patch-vq:'patching up'the video quality problem.
\newblock In \emph{Proceedings of the IEEE/CVF conference on computer vision
  and pattern recognition}, pages 14019--14029, 2021.

\bibitem[Zeng et~al.(2019)Zeng, Li, Cao, and Zhang]{zeng2019reliable}
Hui Zeng, Lida Li, Zisheng Cao, and Lei Zhang.
\newblock Reliable and efficient image cropping: A grid anchor based approach.
\newblock In \emph{Proceedings of the IEEE/CVF conference on computer vision
  and pattern recognition}, pages 5949--5957, 2019.

\bibitem[Zhang et~al.(2018)Zhang, Isola, Efros, Shechtman, and
  Wang]{zhang2018unreasonable}
Richard Zhang, Phillip Isola, Alexei~A Efros, Eli Shechtman, and Oliver Wang.
\newblock The unreasonable effectiveness of deep features as a perceptual
  metric.
\newblock In \emph{Proceedings of the IEEE conference on computer vision and
  pattern recognition}, pages 586--595, 2018.

\bibitem[Zhang et~al.(2025)Zhang, Kou, Wang, Li, Sun, Wang, Li, Wang, Cao, Min,
  et~al.]{zhang2025q}
Zicheng Zhang, Tengchuan Kou, Shushi Wang, Chunyi Li, Wei Sun, Wei Wang, Xiaoyu
  Li, Zongyu Wang, Xuezhi Cao, Xiongkuo Min, et~al.
\newblock Q-eval-100k: Evaluating visual quality and alignment level for
  text-to-vision content.
\newblock In \emph{Proceedings of the Computer Vision and Pattern Recognition
  Conference}, pages 10621--10631, 2025.

\bibitem[Zhou et~al.(2024)Zhou, Wang, Lin, Su, Chen, Tao, Zheng, Yuan, Wan, and
  Zhang]{zhou2024uniaa}
Zhaokun Zhou, Qiulin Wang, Bin Lin, Yiwei Su, Rui Chen, Xin Tao, Amin Zheng, Li
  Yuan, Pengfei Wan, and Di Zhang.
\newblock Uniaa: A unified multi-modal image aesthetic assessment baseline and
  benchmark.
\newblock \emph{arXiv preprint arXiv:2404.09619}, 2024.

\end{thebibliography}
